\documentclass[sigconf]{acmart}
\renewcommand\footnotetextcopyrightpermission[1]{} 
\AtBeginDocument{%
  }
\usepackage{amsmath}
\usepackage{xspace}
\usepackage{hyperref}
\usepackage{subcaption}
\usepackage{url}
\usepackage[table]{xcolor}
\usepackage{booktabs}
\definecolor{darkgreen}{RGB}{1,90,10}
\newcommand{\toolname}{\textsc{Coinvisor}\xspace}
\begin{document}

\title{\toolname: An RL-Enhanced Chatbot Agent for Interactive Cryptocurrency Investment Analysis}



\author{Chong Chen}
\email{chench578@gmail.com}
\affiliation{%
  \institution{Zhejiang University}
  \city{Hangzhou}
  \country{China}
}

\author{Ze Liu}
\email{202422280611@std.uestc.edu.cn}
\affiliation{%
  \institution{University of Electronic Science and Technology of China}
  \city{Chengdu}
  \country{China}
}

\author{Lingfeng Bao}
\email{lingfengbao@zju.edu.cn}
\affiliation{%
  \institution{Zhejiang University}
  \city{Hangzhou}
  \country{China}
}

\author{Yanlin Wang}
\email{wangylin36@mail.sysu.edu.cn}
\affiliation{%
  \institution{Sun Yat-sen University}
  \city{Zhuhai}
  \country{China}
}

\author{Ting Chen}
\email{brokendragon@uestc.edu.cn}
\affiliation{%
  \institution{University of Electronic Science and Technology of China}
  \city{Chengdu}
  \country{China}
}

\author{Daoyuan Wu}
\email{daoyuanwu@In.edu.hk}
\affiliation{%
  \institution{Lingnan University}
  \country{Hong Kong}
}

\author{Jiachi Chen}
\email{chenjch86@mail.sysu.edu.cn}
\affiliation{%
  \institution{Zhejiang University}
  \country{Hangzhou}
}







\begin{abstract}

The cryptocurrency market offers significant investment opportunities, but it also presents challenges, including high price volatility and fragmented information scattered across various websites. Web data integration and analysis have become indispensable for informed cryptocurrency investment decisions. Currently, investors rely on three main approaches for cryptocurrency investment analysis: (1) Manual analysis across various web sources, which depends heavily on individual experience and is time-consuming and prone to bias; (2) Web-based data aggregation platforms, which provide some level of information integration but are often limited in functionality and offer little in-depth analysis; (3) Large language model-based agents, which are typically based on static pretrained models, lacking real-time web data integration, and are limited in performing multi-step reasoning and interactive analysis. To address these challenges, we present \toolname, a reinforcement learning-based web chatbot that provides comprehensive analytical support for cryptocurrency investment decision-making through a multi-agent framework. \toolname integrates diverse web-based analytical capabilities through specialized tools. Its key innovation is a reinforcement learning–based tool selection mechanism that enables multi-step planning and flexible integration of heterogeneous web data sources. This design supports real-time interaction and adaptive analysis of dynamic web content, providing accurate and actionable investment insights. We evaluated \toolname through automated benchmarks on tool calling accuracy and user studies involving 20 cryptocurrency investors who interacted via our web interface. Experiments show that \toolname improves recall by 40.7\% and F1 score by 26.6\% over the base model in web tool orchestration. Additionally, user studies report high satisfaction (4.64/5), with participants preferring \toolname to both general LLMs and existing web-based crypto platforms (4.62/5). We have developed \toolname into a web application with a demonstration available, and will release the complete system as open-source software upon acceptance.

\end{abstract}





\maketitle

\section{Introduction}
Cryptocurrencies have become an important asset class that attracts both individual and institutional investors~\cite{phillip2018new}. While the market offers high potential returns, it also suffers from extreme price volatility, complex market dynamics, and overwhelming amounts of data from multiple sources~\cite{hardle2020understanding}. This combination of high volatility and information overload makes investment decisions difficult and risky. Investors require more effective analytical tools to understand market trends and make informed decisions.

Despite the cryptocurrency market's sophistication, with its requirement for real-time multi-source data analysis and specialized technical knowledge, investors face several critical limitations. A primary challenge is the information gap: conventional large language models (LLMs) cannot access real-time data, while manual monitoring across live feeds is prohibitively labor-intensive. This issue is compounded by severe data fragmentation, forcing investors to inefficiently synthesize multi-dimensional information (e.g., market trends, on-chain activity, and project fundamentals) from numerous disparate platforms. Finally, the specialized knowledge required to interpret this complex data, including an understanding of blockchain technology, smart contract mechanics, tokenomics, and on-chain metrics, poses substantial entry barriers. Novice investors must master these technical domains before making informed decisions.

Current investment tools fall into three categories: traditional financial terminals, crypto-native platforms, and AI-based agents, each with significant limitations. Traditional financial terminals suffer from prohibitive costs and a lack of cryptocurrency-specific functionality. 
Systems like Bloomberg~\cite{RefinitivEikon} and Refinitiv Eikon~\cite{RefinitivEikon} were designed for conventional stock markets, making them technically inadequate for evaluating cryptocurrencies, as they cannot analyze on-chain data sources such as transactions and project backgrounds that are essential to assessing crypto assets. Crypto-native platforms are severely fragmented and limited in analytical depth. While data aggregators such as CoinGecko~\cite{coingecko} provide relatively multi-dimensional information, including real-time prices and latest news, they offer only surface-level data without deep analytical capabilities, such as trend prediction based on candlestick chart patterns or market impact assessment derived from news sentiment analysis.
AI-based agents remain constrained by static knowledge bases and inflexible reasoning architectures. Recent academic systems and LLMs with tool-calling capabilities cannot adapt to rapidly evolving market conditions, as they rely on pre-trained knowledge and rigid operational frameworks~\cite{yu2024fincon,luo2025llm,xiao2024tradingagents}. More critically, these systems lack cryptocurrency-specific analytical tools and are limited to single-step decision-making, producing superficial analyses insufficient for complex investment queries.

To address these challenges, we propose \toolname, an \textbf{\textit{interactive chatbot}} that provides real-time cryptocurrency investment analysis and decision support through natural conversation. \toolname incorporates a comprehensive suite of analytical tools and agents that enable real-time interaction with the rapidly evolving cryptocurrency market, combining multiple data sources and analytical methods to deliver holistic investment insights (refer to Table~\ref{tab:functions}). This design allows users to engage in natural conversation while the system handles complex data retrieval and analysis tasks behind the scenes. Beyond equipping the chatbot with diverse analytical capabilities, we employ reinforcement learning techniques to train the model's multi-step decision-making capability for tool selection. This training enables the chatbot to autonomously determine and invoke the most relevant tools based on user queries, ensuring contextually appropriate and comprehensive responses. Through this interactive dialogue interface, \toolname autonomously gathers, processes, and synthesizes multi-dimensional information to deliver personalized and actionable investment insights.


We evaluated the reinforcement learning-enhanced tool selection mechanism using a manually annotated dataset of 500 cryptocurrency investment queries, measuring the model's F1 score in selecting appropriate tools for each query. Additionally, we conducted user studies where 20 cryptocurrency investors interacted with the system through real investment scenarios and evaluated its performance through structured questionnaires. Our evaluation shows that \toolname substantially enhances both analytical capability and user experience. The reinforcement learning–enhanced caller model improved recall by 40.7\% and F1 score by 26.6\%, while maintaining high precision. In user studies with 20 participants, \toolname consistently received high satisfaction scores across all core functionalities (4.64/5), and was preferred over both general-purpose LLMs and crypto-specific platforms (4.62/5). Participants highlighted its time-saving efficiency (95\%), beginner-friendly design (80\%), and clearer analytical logic (60\%), underscoring its advantages in providing accurate and accessible investment support.

In summary, this work makes the following contributions:
\begin{itemize}
    \item We present \toolname, a novel chatbot system tailored for cryptocurrency investment, which unifies diverse data sources and supports interactive decision making.
    \item We develop a reinforcement learning–based caller model for multi-step tool selection, combined with domain-specific tool design and a multi-agent framework, enabling adaptive and coordinated investment analysis.
    \item We have developed \toolname into a web application that will be publicly released upon paper acceptance to enable practical deployment and user access.
    \item The demo video of \toolname, the training/evaluation datasets, and trained model weights are at \url{https://github.com/BugmakerCC/Coinvisor}.
\end{itemize}
\section{Background}

\subsection{Cryptocurrency Investment}

The decision-making process for cryptocurrency investment is multidimensional and complex~\cite{zhao2021financial,inci2019cryptocurrencies,almeida2022systematic}. Traditional financial analysis frameworks are inadequate for addressing this complexity~\cite{lee2017cryptocurrency,morck1990stock,yuniningsih2017analysis}. Investors must assess projects through technical fundamentals (e.g., tokenomics, utility, and development activity), market dynamics (e.g., trading volume, liquidity, and price trends), and qualitative factors (e.g., team credibility, community engagement)~\cite{luo2025llm}. The absence of standardized valuation models necessitates that investors synthesize information from diverse sources, including exchange data~\cite{binance}, blockchain explorers~\cite{etherscan}, developer repositories~\cite{github}, and others, to formulate informed investment judgments. However, there are no tools that can consolidate all these different types of data, process them, and provide practical investment advice. This creates a significant gap between the information we have and the information we need to make good decisions.

\begin{figure*}[h]
\centering
\includegraphics[width=\linewidth]{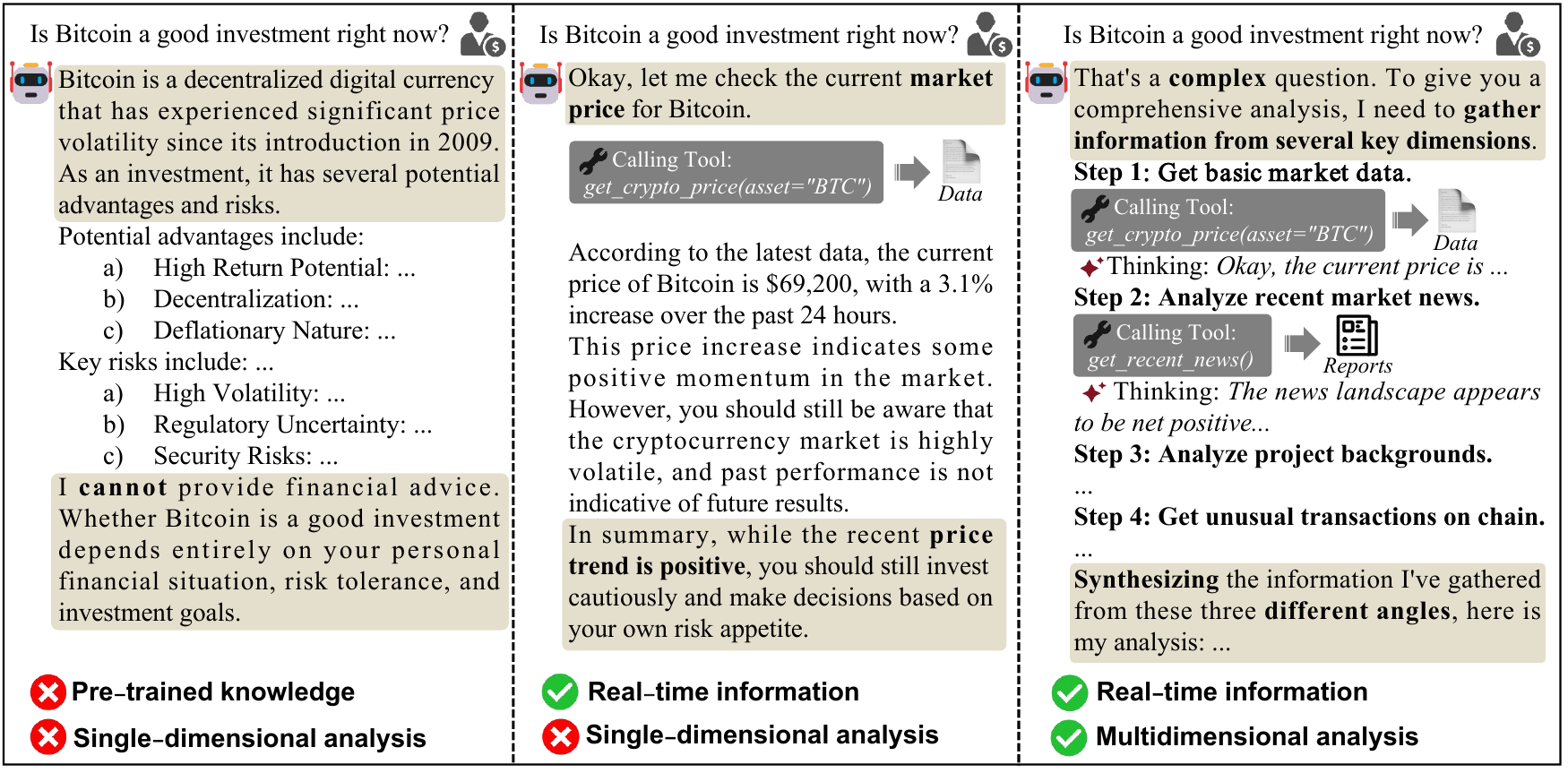}
\caption{Comparison of three approaches to cryptocurrency investment queries: (left) standard LLM relying on pre-trained knowledge, (middle) tool-augmented LLM lacking multi-dimensional analysis capability with minimal tool invocation, and (right) multi-agent approach with coordinated multi-tool orchestration for comprehensive multidimensional analysis.}
\label{fig:case}
\end{figure*}

\subsection{LLM-based Approaches}

\noindent\textbf{Background.}
The emergence of LLMs, such as GPT~\cite{achiam2023gpt}, Llama~\cite{grattafiori2024llama}, and DeepSeek~\cite{liu2024deepseek}, has enabled their use as reasoning engines for autonomous agents, showing potential in complex domains like finance by decomposing high-level goals into actionable plans. While real-time information can be obtained via web search~\cite{websearch} or Retrieval-Augmented Generation (RAG)~\cite{lewis2020retrieval,gao2023retrieval,zhao2024retrieval,yu2022retrieval}, this method is often inefficient for the frequent, structured queries common in financial analysis, such as fetching a specific asset's price. Direct tool calling~\cite{shen2024llm,yuan2024easytool,wang2024toolflow} thus emerges as the most viable option. However, a prevalent design philosophy in LLM finetuning aims to prevent tool overuse and maximize efficiency by intentionally training these models to resolve user prompts with the fewest tool calls possible~\cite{xu2024reducing,ross2025when2call}.

\noindent\textbf{Motivating Example.}
To illustrate the limitations of current LLM-based approaches in cryptocurrency investment, we present a motivating example in Figure~\ref{fig:case}, which compares three different responses to the query "Is Bitcoin a good investment right now?" A standard LLM, acting as a standalone reasoning agent, is fundamentally constrained by its static, pre-trained knowledge (Figure~\ref{fig:case}, left). This core limitation renders it incapable of accessing real-time information. Consequently, it can only provide generic, outdated information, making it inherently insufficient for analyzing the volatile cryptocurrency market. As shown in the middle panel of Figure~\ref{fig:case}, tool-augmented LLMs lead to a different problem: a flawed, single-dimensional analysis. A typical agent might fetch the current price but will proceed no further. This efficiency-driven behavior directly conflicts with the demands of cryptocurrency investment. An informed decision requires a comprehensive, multidimensional analysis, as depicted in the ideal scenario on the right of Figure~\ref{fig:case}. Such an analysis necessitates the systematic coordination of multiple tools. The critical gap, therefore, lies between the agent's efficiency-driven design and the domain's requirement for thorough, multi-tool information retrieval, where comprehensive coverage is prioritized over minimal tool invocation to avoid misleading conclusions.

\section{Method}

\begin{figure*}[h]
\centering
\includegraphics[width=\linewidth]{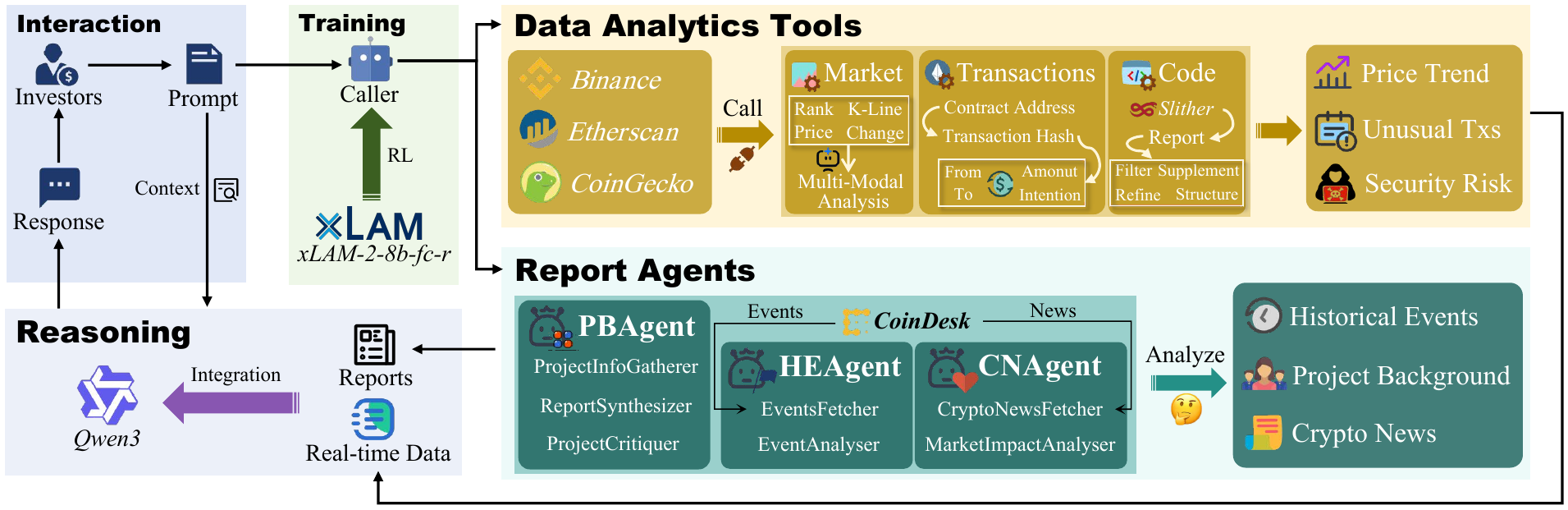}
\caption{The overview of \toolname.}
\label{fig:overview}
\end{figure*}

\toolname is a reinforcement learning–enhanced chatbot system designed to support cryptocurrency investment decisions through multi-step tool usage and planning. As shown in Figure~\ref{fig:overview}, given an investor's prompt, the system initiates a caller module that interacts with an \textbf{RL-tuned LLM} (\textit{xLAM-2-8b-fc-r}~\cite{prabhakar2025apigen}), which iteratively selects and invokes a sequence of tools in a multi-step decision-making process. These tools fall into two categories: \textbf{Data Analytics Tools} and \textbf{Report Agents}. Data analytics tools integrate external APIs (e.g., \textit{Binance}~\cite{binance}, \textit{Etherscan}~\cite{etherscan}, \textit{CoinGecko}~\cite{coingecko}) to retrieve real-time data. This enables the system to conduct analysis on aspects such as price trends, unusual transactions, and potential security risks. Report agents activate specialized sub-agents, i.e., PBAgent, HEAgent, and CNAgent, which are responsible for gathering and analyzing project background information, historical events, and crypto-related news, respectively. 
The tool outputs are integrated and processed by the reasoning model (Qwen3~\cite{qwen3}), which synthesizes the final response based on the user's original query and the collected information. This interactive framework enables users to engage in continuous dialogue with the system to obtain optimal responses.
\vspace{-0.3cm}

\subsection{Tool Learning}
To enhance our system's ability to reason and plan, we move beyond supervised fine-tuning and employ Reinforcement Learning (RL) to train its multi-step tool-call decision-making capabilities. The primary objective is to teach the model not just to call a single correct tool, but to autonomously generate a sequence of tool calls that collectively gather comprehensive information to address a user's complex query. We formulate this as a sequential decision-making problem and utilize the Proximal Policy Optimization (PPO) algorithm~\cite{schulman2017proximal} to optimize the agent's policy.

\subsubsection{Problem Formulation}

We model the tool-call planning task as a finite-horizon Markov Decision Process (MDP)~\cite{puterman1990markov}, defined by the tuple $(\mathcal{S}, \mathcal{A}, \mathcal{P}, \mathcal{R}, \gamma)$:

\begin{itemize}
    \item \textbf{State} ($S_t \in \mathcal{S}$): A state at timestep $t$ is the conversational history up to that point. It is represented as a sequence of turns, $S_t = (c_0, c_1, \ldots, c_t)$, where each $c_i$ includes the role (system, user, assistant, or tool) and content. The initial state $S_0$ consists of the system prompt (containing tool definitions) and the user's initial query.
    
    \item \textbf{Action} ($A_t \in \mathcal{A}$): The action space $\mathcal{A}$ is the model's vocabulary space. An action $A_t$ is a text sequence $y$ generated by the policy network, parsed into a structured list of tool calls $T_t = \texttt{parse}(y)$. The agent can choose any combination of tool calls or terminate by producing the special \texttt{no\_tool\_call} action.
    
    \item \textbf{Policy} ($\pi_\theta$): The policy is a language model parameterized by $\theta$, generating actions stochastically: $A_t \sim \pi_\theta(\cdot | S_t)$. It is optimized to maximize the expected cumulative reward.
    
    \item \textbf{Transition} ($\mathcal{P}$): Transitions are deterministic. After taking action $A_t$ in state $S_t$, the next state is $S_{t+1} = S_t \oplus A_t \oplus \texttt{Simulate}(T_t)$, where $\oplus$ denotes concatenation, and \texttt{Simulate} returns confirmation messages for the executed tools.
    
    \item \textbf{Episode and Trajectory} ($\tau$): An episode corresponds to a full interaction process initiated by a user query and terminated by a stop action or a predefined maximum step limit. It consists of a sequence of state-action pairs, forming a trajectory $\tau = (S_0, A_0, \ldots, S_T)$, which is used for reward computation and policy optimization.
\end{itemize}

\subsubsection{Reward Function Design}
\label{subsubsec:rewardfunction}
We design a hybrid reward function that assigns a terminal reward $R(\tau)$ to each trajectory $\tau$ based on syntactic correctness and semantic quality. The complete tool call history for an episode is denoted $H_\tau = \{T_0, T_1, \ldots, T_{T-1}\}$.

\begin{equation}
R(\tau) = w_{\text{judge}} \cdot R_{\text{judge}}(H_\tau, Q) + w_{\text{correct}} \cdot R_{\text{correct}}(H_\tau)
\end{equation}

where $Q$ is the user query, $H_\tau = \{T_0, T_1, ..., T_{T-1}\}$ is the complete tool call history, and weights are set as $w_{judge} = 0.7$, $w_{correct} = 0.3$. We prioritize semantic quality ($R_{judge}$) over syntactic correctness ($R_{correct}$) since our base model already demonstrates high proficiency in generating syntactically valid tool calls.

\paragraph{Syntactic Correctness Reward ($R_{\text{correct}}$):}

This score checks schema adherence using rule-based checks: 1) \textit{Fatal Errors}: Invalid tool names or unparsable JSON result in a score of 0, 2) \textit{Major Errors}: Missing or invalid parameters, wrong data types, and 3) \textit{Redundancy}: Duplicate tool calls with the same arguments.
We define:
\begin{equation}
R_{\text{correct}}(H_\tau) = \max\left(0, \frac{C_{\max} - \sum_{v \in V} p_v}{C_{\max}}\right)
\end{equation}

\noindent where $C_{\max} = 3.0$, $V$ is the set of violations, and $p_v$ is the penalty of violation $v$.

\paragraph{Semantic Quality Reward ($R_{\text{judge}}$):}
This component evaluates the strategic quality of the chosen tools. We employ an external, powerful Large Language Model (LLM) serving as a judge to provide a nuanced, semantic reward signal. The evaluation is not based on a simple matching task but is guided by a meticulously engineered prompt that enables the judge to perform sophisticated reasoning about the model's tool-call plan.

The evaluation process of the judge model is driven by a structured prompt (refer to Appendix~\ref{app:judge_prompt}) that explicitly defines the model's role, objectives, evaluation criteria, and expected output format. As shown in Figure~\ref{fig:prompt}, the system prompt instructs the model to act as an LLM-as-a-Judge for a tool-augmented AI assistant operating in the cryptocurrency domain. The judge is tasked with assessing the quality of a tool call plan generated in response to a user query. It is provided with the complete list of available [TOOL LIST], ensuring full awareness of the assistant’s action space. The core of the evaluation rests on two explicit criteria, for which the judge must provide scores on a floating-point scale from 0 to 10:
\begin{itemize}
    \item \textit{Information Coverage} ($S_{\text{cov}}$): This metric assesses the breadth and completeness of the model's plan. It is designed to reward the model for formulating a comprehensive investigation strategy that covers multiple relevant information dimensions, such as market data, on-chain analysis, news, and project background. The underlying principle is to incentivize the selection of a diverse set of tools, thereby encouraging multi-faceted analysis for complex queries over narrow, one-dimensional approaches.
    
    \item \textit{Relevance} ($S_{\text{rel}}$): This metric assesses the precision of the plan. It evaluates whether the chosen tools and their arguments are directly appropriate and aligned with the user's specific query, penalizing irrelevant or misused tools.
\end{itemize}

For each trajectory $\tau$, the judge receives the initial user query $Q$ and the complete tool call history $H_{\tau}$. It then generates a structured JSON response containing the scores $S_{\text{cov}}$ and $S_{\text{rel}}$. These scores are combined into the final semantic reward:

\begin{equation}
R_{\text{judge}}(H_\tau, Q) = \frac{1}{10} \left( w_{\text{cov}} \cdot S_{\text{cov}} + w_{\text{rel}} \cdot S_{\text{rel}} \right)
\end{equation}


The preceding $\frac{1}{10}$ coefficient normalizes the weighted sum to a standardized range. We set weights to $w_{\text{cov}} = 0.8$ and $w_{\text{rel}} = 0.2$ based on empirical analysis. The base policy, due to prior supervised fine-tuning, consistently generates highly relevant tool calls with minimal variance in $S_{\text{rel}}$ scores, providing a weak discriminative signal. In contrast, $S_{\text{cov}}$ exhibits substantial variance across plans, effectively distinguishing comprehensive from minimal strategies. Thus, we prioritize coverage to focus reinforcement learning on improving strategic breadth, while the smaller relevance weight maintains existing capabilities as a regularizing constraint.


\subsubsection{PPO Training and Optimization}
\begin{figure}[h]
\centering
\includegraphics[width=\linewidth]{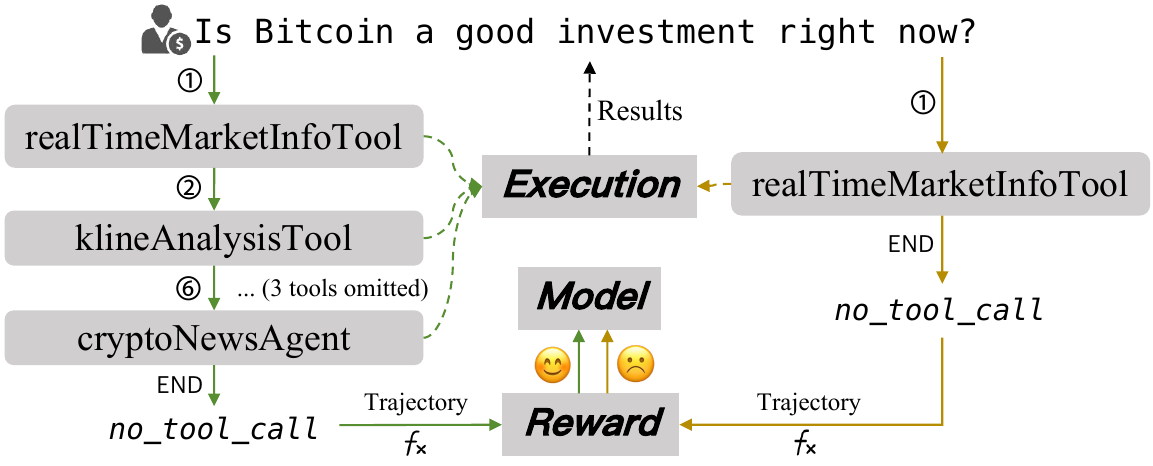}
\caption{Examples of different trajectories during model training. The ``\textcolor{green!50!black}{$\rightarrow$}'' represents a high-reward trajectory, where the model generates a comprehensive sequence of tool calls. In contrast, the ``\textcolor{yellow!70!black}{$\rightarrow$}'' represents a low-reward trajectory, involving only a minimal number of tool calls and limited information gathering.}
\label{fig:train}
\end{figure}

We constructed a dataset of 1,000 cryptocurrency investment-related prompts using \textit{Gemini 2.5 Pro}~\cite{gemini}. To prevent overfitting, the dataset includes diverse prompt categories, such as real-time information queries, investment advice, and knowledge-based Q\&A, each requiring different tool call strategies. We employed \textit{LoRA}~\cite{hu2022lora} for training, as it helps maintain the stability of the base model. Figure~\ref{fig:train} illustrates the parameter update process during model training. Given a prompt, the model generates a sequence of tool calls, with each tool execution simulated using the model’s predicted parameters. To reduce training overhead, we use simulated execution results instead of real tool outputs, as actual execution is costly due to nested agents, often yields long outputs, and is unnecessary for a model focused solely on decision-making rather than result interpretation. These simulated results are then fed back into the model as part of the input state, enabling it to iteratively generate the next tool call based on the evolving context. This process continues until the model stops generating tool calls and terminates. After that, all generated tool call sequences, including tool names, parameter names, and parameter types, are evaluated by the reward function (refer to Section~\ref{subsubsec:rewardfunction}). The resulting reward value is then used to update the model parameters. We use PPO to optimize $\pi_\theta$. The objective balances advantage maximization and KL regularization:

\begin{equation}
\max_\theta \,\, \mathbb{E}_{(S, A) \sim \mathcal{D}} \left[ \pi_\theta(A|S) \cdot \hat{A}^{\pi_{\text{old}}}(S, A) - \beta \cdot \text{KL}\left[ \pi_{\text{old}}(\cdot|S) \,\|\, \pi_\theta(\cdot|S) \right] \right]
\end{equation}

where:
\begin{itemize}
    \item $\pi_{\text{old}}$ is the frozen reference policy,
    \item $\hat{A}^{\pi_{\text{old}}}$ is the estimated advantage,
    \item $\text{KL}$ is the Kullback-Leibler divergence,
    \item $\beta$ is the KL penalty coefficient (dynamically adjusted).
\end{itemize}

The critic, parameterized by $\phi$, minimizes value loss:

\begin{equation}
\min_\phi \,\, \mathbb{E}_{(S, R_{\text{ep}}) \sim \mathcal{D}} \left[ \left( V_\phi(S) - R_{\text{ep}} \right)^2 \right]
\end{equation}

\subsection{Data Analytics Tools}
To support comprehensive cryptocurrency analysis, \toolname incorporates structured functions that extract and analyze information across market dynamics, on-chain activity, and smart contract security. Each function retrieves objective data and applies domain-specific techniques to generate actionable insights (refer to Appendix~\ref{app:detailed_functions} for more detailed functional descriptions).

\paragraph{Market Analysis.}
The market analysis function retrieves real-time information by calling external APIs such as \textit{Binance}~\cite{binance}. When the user specifies a particular token, the function can return detailed market data, including the current price, price change rate, trading volume, and K-line (candlestick) chart data. The inclusion of visual information introduces a multimodal aspect to the analysis. In cases where the user’s investment target is unclear, the function can instead fetch ranked market overviews, such as top gainers, losers, and popular tokens with higher trading volume. These structured outputs provide a comprehensive snapshot of market conditions. 

\paragraph{Transaction Analysis.}
Large-scale token transfers often signal significant market movements. The transaction analysis function traces these movements by first querying CoinGecko~\cite{coingecko} to retrieve the contract address for a given token name, then utilizing the Etherscan API~\cite{etherscan} to collect recent on-chain transactions. The function filters transactions below \$1,000,000 to focus on significant transfers, then analyzes sender and receiver addresses to infer transaction intent (e.g., whale accumulation, exchange deposits, and inter-wallet transfers). Structured data entries that capture transaction value, timestamp, directionality, and inferred intent are returned to facilitate accurate decision-making.

\begin{figure}[h]
\centering
\includegraphics[width=\linewidth]{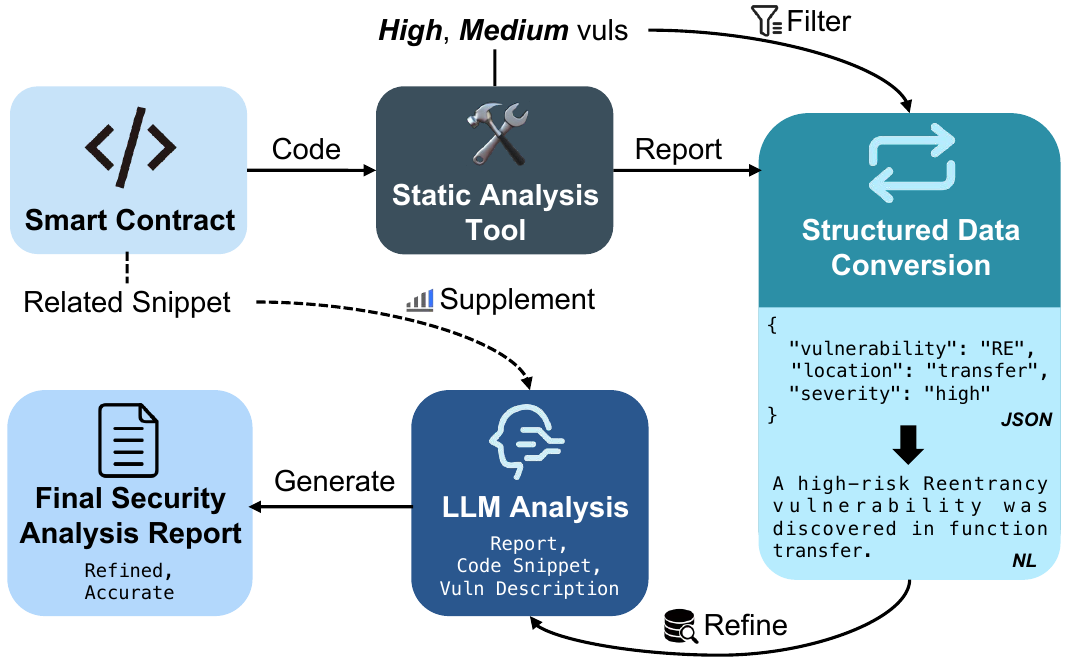}
\caption{The process of conducting security analysis on the source code of smart contracts.}
\label{fig:code_analysis}
\end{figure}

\paragraph{Code Analysis.}
Smart contract security is a crucial factor in cryptocurrency investment, as vulnerabilities in contract logic can result in catastrophic asset loss. The code analysis function implements a multi-stage security assessment pipeline as illustrated in Figure~\ref{fig:code_analysis}. Given a token identifier, the system first retrieves the verified contract source code from Etherscan~\cite{etherscan}, then applies the static analysis tool Slither~\cite{feist2019slither} to detect known vulnerability patterns. The initial report is filtered to retain only high- and medium-severity issues, eliminating noise from low-impact findings. Subsequently, an LLM processes this technical output to identify potential false positives and generate a refined natural language summary that highlights vulnerability types, affected code locations, and severity levels in a context relevant to investment. 

\subsection{Report Agents}
\begin{figure}[h]
\centering
\includegraphics[width=\linewidth]{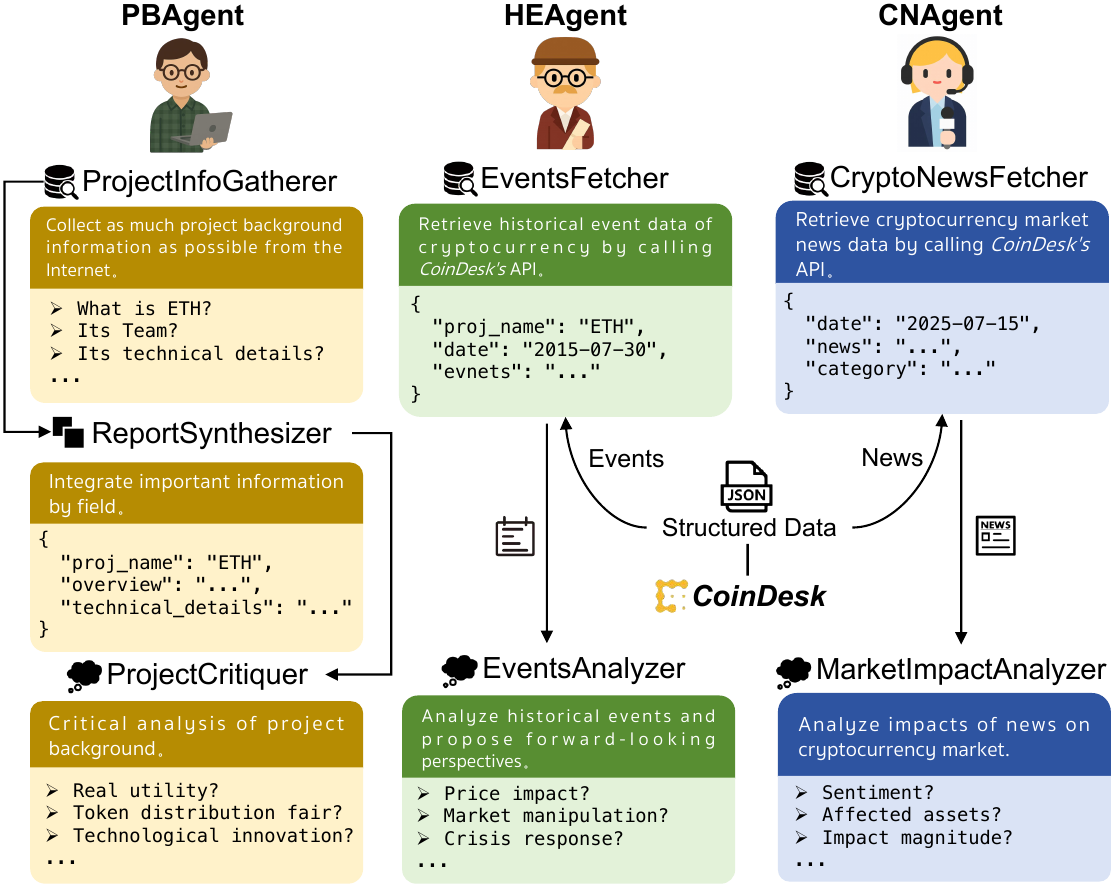}
\caption{The workflow of multiple agents integrated in \toolname.}
\label{fig:agents}
\end{figure}

In addition to function calls, \toolname\ employs a set of role-specialized agents to retrieve and interpret contextual information that cannot be directly obtained through APIs. Each agent is implemented as a dedicated LLM prompt template, prompting the model to act as a domain-specific assistant that gathers and analyzes textual knowledge. These agents operate in parallel and contribute complementary insights that support more informed investment reasoning. As illustrated in Figure~\ref{fig:agents}, \toolname\ uses three such agents: \textbf{PBAgent}, \textbf{HEAgent}, and \textbf{CNAgent} (refer to Appendix~\ref{app:detailed_functions} for more detailed descriptions).

\textbf{PBAgent} (Project Background Agent) operates through three sequential components: \textit{ProjectInfoGatherer} issues broad web queries to collect project information from official websites, GitHub repositories, and whitepapers; \textit{ReportSynthesizer} structures this unstructured content into fields including project overview, founding team, and technical specifications; and \textit{ProjectCritiquer} performs critical analysis evaluating real utility, token distribution fairness, and technological innovation to assess fundamental investment value. \textbf{HEAgent} (Historical Events Agent) comprises two modules: \textit{EventsFetcher} retrieves historical cryptocurrency events from \textit{CoinDesk} API~\cite{coindesk} with structured date and event data, and \textit{EventsAnalyzer} evaluates price impact, potential market manipulation, and crisis response patterns to identify cause-and-effect relationships and long-term risk patterns. \textbf{CNAgent} (Crypto News Agent) similarly contains two components: \textit{CryptoNewsFetcher} queries \textit{CoinDesk} API~\cite{coindesk} for real-time market news with date and category metadata, and \textit{MarketImpactAnalyzer} assesses sentiment, affected assets, and impact magnitude to generate forward-looking market insights. These agents coordinate through structured data exchange, with each sub-module performing specific roles in the information gathering and analysis pipeline, ultimately synthesizing comprehensive investment intelligence that complements quantitative data from analytical functions.

\section{Evaluation}
In this section, we evaluate the performance of \toolname from two perspectives. First, we examine the effectiveness of the reinforcement learning–based tool call mechanism (RQ1) by evaluating the caller model on a synthesized and annotated dataset of cryptocurrency investment prompts. Second, we assess the practical utility of the overall \toolname system (RQ2) through a user study involving both novice and experienced cryptocurrency investors. These two evaluations collectively demonstrate \toolname's capabilities in both accurate tool call and real-world decision support.

\subsection{RQ1: Performance of the Caller model}


\noindent \textbf{Dataset.} We constructed a synthetic evaluation dataset of 500 cryptocurrency investment prompts using \textit{Gemini 2.5 Pro}~\cite{gemini}. We guided the generation to create diverse query types, ranging from simple information requests to complex investment advice, ensuring coverage of varied tool-call requirements, including single API calls and multi-step reasoning plans that coordinate multiple tools. To ensure evaluation validity, we maintained strict separation between the evaluation and training datasets. Each prompt was manually annotated by two domain experts with an ideal tool-call sequence representing the most comprehensive, accurate, and efficient actions to address the query. Annotations matched the model's output format. The experts annotated independently, then cross-checked results and resolved all discrepancies through discussion to establish unified ground-truth labels.

\noindent \textbf{Experimental Setup.} We adopt \textit{Llama-xLAM-2-8b-fc-r}~\cite{prabhakar2025apigen} as the base model, which has been fine-tuned specifically for tool-calling tasks and ranks four on the \textit{Berkeley Function-Calling Leaderboard V3}~\cite{patil2025bfcl}. Among all models on the leaderboard, it is outperformed only by three significantly larger variants (32B and 70B), making the 8B version a favorable trade-off between tool-calling performance and computational cost. \textit{QLoRA} was used for parameter-efficient fine-tuning. The model was loaded in 4-bit precision with a bfloat16 compute data type, and the \textit{LoRA} configuration specified a rank of 8 and an alpha of 32. The model was trained for three epochs using the Proximal Policy Optimization (PPO) algorithm from the \textit{trl} library~\cite{vonwerra2022trl}. Key hyperparameters included a learning rate of $5 \times 10^{-6}$, a batch size of 8, and an adaptive KL controller with a target of 0.01. To encourage exploration, actions during training were generated via stochastic sampling (\texttt{do\_sample=True}), with a temperature of 0.7 and \texttt{top\_p} of 0.8. Moreover, we selected \textit{Qwen3-30B-A3B}~\cite{qwen3technicalreport} as the judge model due to its recent release and moderate parameter count, which ensured good inference performance. For reasoning and analysis tasks following tool execution, including function result interpretation and agent-based analysis, we employed Qwen3 (qwen-plus)~\cite{qwen3} as the primary language model across all system components.

\noindent For evaluation, the trained LoRA adapter was loaded onto the quantized base model with greedy decoding (\texttt{do\_sample=False}) to ensure reproducibility. The model performed up to 6 sequential tool-calling decisions with settings consistent with those used during training. Experiments were conducted on an Ubuntu server with eight NVIDIA H100 80GB GPUs.

\noindent \textbf{Results.} We quantify the performance of our caller model using Precision, Recall, and F1 Score. For each prompt in our evaluation dataset, let $G$ be the set of tool calls generated by the model and $T$ be the set of ground-truth tool calls from our manual annotations. The Precision ($P$) and Recall ($R$) for a single prompt are defined as: $P = \frac{|G \cap T|}{|G|}, \quad R = \frac{|G \cap T|}{|T|}$. A special case is handled for prompts where no tool call is required ($T = \emptyset$): if the model also generates no tools ($G = \emptyset$), both $P$ and $R$ are set to 1.0; if the model incorrectly invokes any tool ($G \neq \emptyset$), they are set to 0.0. To calculate the final metrics, we first compute the average Precision ($\bar{P}$) and average Recall ($\bar{R}$) across all prompts in the dataset. The overall F1 Score is then calculated using these averaged values: $F_1 = 2 \cdot \frac{\bar{P} \cdot \bar{R}}{\bar{P} + \bar{R}}$.

\noindent The aggregated results of this evaluation are presented in Table~\ref{tab:model_performance} and show a marked improvement in the performance of the caller model compared to the base model. The most significant gain is in Recall, where our caller model achieves a score of 0.390, a substantial \textbf{40.7\% increase} over the base model's 0.277. This indicates that the reinforcement learning process has successfully encouraged a more comprehensive and exploratory tool-calling strategy. While this exploratory approach leads to a slight decrease in Precision, we consider this an acceptable trade-off in the cryptocurrency investment domain, where access to broader information can be beneficial. The overall effectiveness of this trade-off is confirmed by the F1 Score, which harmonizes both metrics and shows a significant \textbf{26.6\% improvement}, rising to 0.528. This substantial gain validates that our training methodology effectively enhances the model's ability to make more useful and complete tool-calling decisions, thus successfully addressing RQ1.


\begin{table}[h]
\centering
\caption{Performance comparison between base model and RL-tuned caller model on the evaluation dataset.}
\label{tab:model_performance}
\begin{tabular}{lccc}
\toprule
\textbf{Model} & \textbf{Recall} & \textbf{Precision} & \textbf{F1 Score} \\
\midrule
Base & 0.277 & 0.844 & 0.417 \\
\rowcolor{blue!8}
RL-tuned & 0.390 \textcolor{darkgreen}{(\textbf{+40.7\%})} & 0.819 \textcolor{red}{(\textbf{-3.0\%})} & 0.528 \textcolor{darkgreen}{(\textbf{+26.6\%})} \\
\bottomrule
\end{tabular}
\end{table}

\subsection{RQ2: Effectiveness of the \toolname system}

To assess the overall effectiveness of \toolname as a complete agent system, we conducted a \textbf{user study} in which participants interacted with the system and then completed a structured questionnaire~\footnote{\url{https://www.wjx.cn/vm/ra0SBUt.aspx\#}}. The user study was designed around several key dimensions: (i) accuracy, coverage, and clarity of outputs across different functional modules (market analysis, project background analysis, etc.); (ii) quality of tool-selection and reasoning processes during task execution; (iii) comparative usefulness against general-purpose LLMs (e.g., \textit{GPT-4o}~\cite{gpt4o}) and conventional crypto data platforms (e.g., \textit{CoinGecko}~\cite{coingecko}); and (iv) overall user experience, including efficiency gains, reduced entry barriers, and perceived investment support. Open-ended questions further solicited qualitative feedback on the system’s strengths, shortcomings, and areas for improvement. 

\paragraph{Participants.}
We recruited a diverse group of 20 participants covering a wide range of roles and experience levels in the cryptocurrency domain. This sample size aligns with established practices in agent design research~\cite{petrov2023dream,zhang2025prompting}. Participants included \textit{novice} investors (with less than one year of involvement), \textit{intermediate} users (1–3 years), and \textit{experienced} participants (3 years or more, including professional analysts and researchers). This stratification ensured that the evaluation reflected both the needs of newcomers seeking accessible guidance and those of advanced users requiring deeper analytical support. 

\paragraph{Procedure.}
Each participant was invited to engage in two sequential tasks. First, they interacted with \toolname in a controlled environment, where they were encouraged to explore the system’s full functionality by posing self-formulated, investment-related queries. These interactions spanned multiple modules of the system. This phase was designed to simulate realistic decision-making scenarios and allow participants to experience the system's tool coordination capabilities in practice. After the interaction phase, participants were required to complete the evaluation questionnaire. The survey systematically covered their perceptions of (i) functional outputs from each module (e.g., accuracy, clarity, comprehensiveness); (ii) system-level reasoning (e.g., appropriateness and efficiency of tool invocation sequences); (iii) comparative usefulness against existing solutions (both general-purpose LLMs and crypto-specific platforms); and (iv) overall experience, including time savings, usability, and reduction of subjective bias.

\paragraph{Measures.}
The questionnaire primarily employed five-point Likert scales~\cite{joshi2015likert} (1 = strongly disagree, 5 = strongly agree, with an additional N/A option), enabling quantitative evaluation across accuracy, comprehensiveness, reasoning clarity, and practical value dimensions. Complementing these structured measures, open-ended questions allowed participants to elaborate on the most helpful aspects of the system, encounter issues, and express further expectations. This mixed design ensured that both numerical trends and nuanced perspectives could be captured.


\paragraph{Results.}
We report the results of the user study using both descriptive statistics and visualizations, thereby capturing participants' overall perceptions of \toolname. The study involved 20 participants with varying cryptocurrency investment experience: 5 participants had less than 1 year of experience, 10 participants had 1–3 years of experience, and 5 participants had over 3 years of experience. Results are based on the 20 valid questionnaires collected and follow a consistent scoring methodology to ensure comparability across different evaluation dimensions. Unless otherwise stated, all Likert-scale~\cite{joshi2015likert} items were rated on a five-point scale, and the reported scores represent the arithmetic mean of all participants' responses, with error bars indicating standard deviations. 


\begin{table}[h]
\centering
\caption{User satisfaction scores across \toolname's core functionalities (N=20, 5-point Likert scale).}
\label{tab:functionality_scores}
\begin{tabular}{lcc}
\toprule
\textbf{Functionality} & \textbf{Average Score} & \textbf{Std. Deviation} \\
\midrule
Transaction & 4.72 & 0.51 \\
Market Information & 4.63 & 0.55 \\
Smart Contract Security & 4.57 & 0.54 \\
Historical Events & 4.63 & 0.56 \\
Project Background & 4.61 & 0.54 \\
Crypto News & 4.66 & 0.53 \\
\midrule
\textbf{Overall Average} & \textbf{4.64} & \textbf{0.54} \\
\bottomrule
\end{tabular}
\end{table}

Table~\ref{tab:functionality_scores} presents a comprehensive evaluation of \toolname's six core functionalities through a radar chart visualization. Each functional dimension was assessed through a matrix-style questionnaire item (Q5-Q10), which contained multiple sub-dimensions that captured different aspects of the user experience. For instance, the ``market information analysis'' dimension (Q6) comprised three distinct sub-dimensions: information accuracy, analytical logic, and presentation clarity. The composite score for each functionality represents the mean value calculated by aggregating all ratings across these sub-dimensions from all 20 participants. 

As visualized in Table~\ref{tab:functionality_scores}, all six functionalities achieved consistently high average scores (4.64) on the 5-point Likert scale. These robust scores demonstrate \toolname's effectiveness across its complete functional spectrum. Furthermore, the remarkably small standard deviations (ranging from 0.51 to 0.56) indicate strong consensus among participants, suggesting that users consistently perceived \toolname as reliable and valuable regardless of their individual backgrounds or usage patterns. This combination of \textbf{high means} and \textbf{low variability} provides compelling evidence for the system's overall quality and user satisfaction.

\begin{figure}[H]
    \centering
    \begin{subfigure}[t]{0.49\linewidth} 
        \centering
        \includegraphics[width=\textwidth]{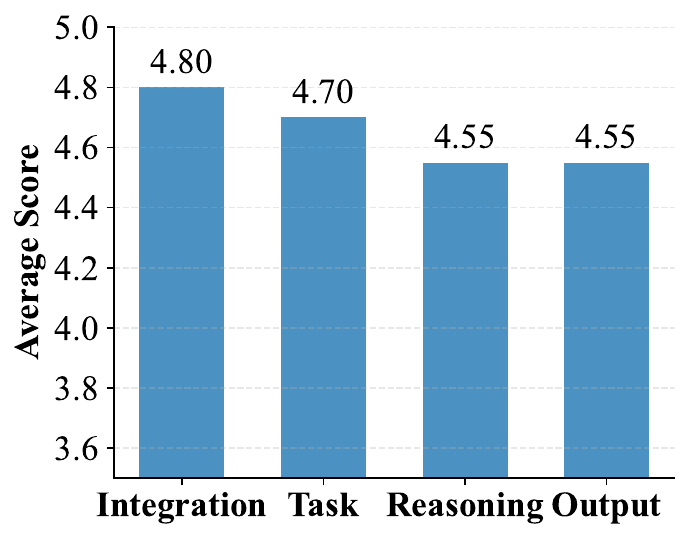} 
        \caption{Comparison between \toolname and general LLMs, e.g., \textit{GPT-4o}.}
        \label{fig:gpt4o}
    \end{subfigure}
    \hfill
    \begin{subfigure}[t]{0.44\linewidth}
        \centering
        \includegraphics[width=\textwidth]{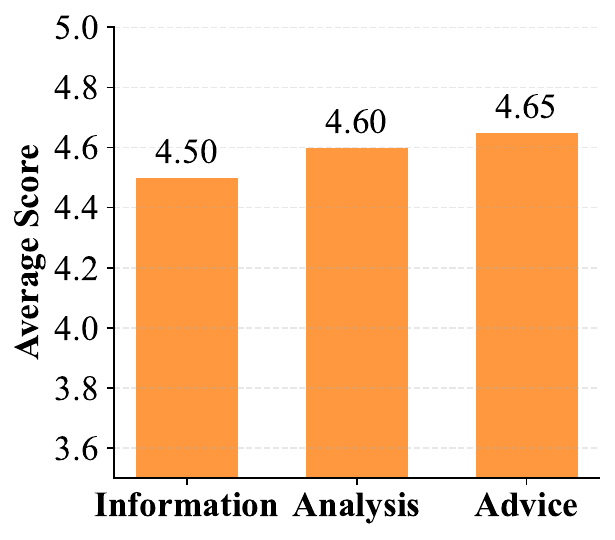}
        \caption{Comparison between \toolname and crypto data platforms, e.g., \textit{CoinGecko}.}
        \label{fig:coingecko}
    \end{subfigure}
    \caption{Comparison between \toolname and baselines.}
    \label{fig:baseline}
\end{figure}

Figure~\ref{fig:baseline} compares \toolname against two baseline categories (Q12-Q13): general LLMs (represented by \textit{GPT-4o}~\cite{gpt4o}), which excel in conversation but lack domain specialization, and specialized cryptocurrency platforms (represented by \textit{CoinGecko}~\cite{coingecko}), which provide comprehensive market data but limited analytical depth and interactivity. \toolname demonstrates superior performance across all dimensions. Compared to general LLMs, it achieves high scores in information integration (4.80), task understanding and decomposition (4.70), reasoning clarity (4.55), and output practicality (4.55). Against specialized platforms, \toolname shows strong performance in information comprehensiveness (4.50), analytical depth (4.60), and investment advice quality (4.65).

\begin{table}[htbp]
\centering
\caption{Perceived Improvements of \toolname Compared to Traditional Analysis Methods.}
\label{tab:user_feedback}
\begin{tabular}{lcc}
\toprule
\textbf{Option} & \textbf{Count (N=20)} & \textbf{Percentage} \\
\midrule
\textit{Information Time Savings} & 19 & 95\% \\
\textit{Reduced Human Bias} & 9 & 45\% \\
\textit{Clearer Analytical Logic} & 12 & 60\% \\
\textit{Lower Expert Costs} & 9 & 45\% \\
\textit{Beginner-Friendly Access} & 16 & 80\% \\
\textit{Personalized System Interaction} & 14 & 70\% \\
\bottomrule
\end{tabular}
\end{table}

As shown in Table~\ref{tab:user_feedback}, user feedback (Q14) indicates that \toolname's most significant advantage over traditional methods lies in information time savings, with 95\% of participants recognizing its efficiency benefits. The system also demonstrates strong accessibility, with 80\% of users noting its beginner-friendly design and 70\% valuing its personalized interaction capabilities. While reduced human bias and cost savings received more moderate recognition (45\% each), these results collectively confirm \toolname's effectiveness in enhancing cryptocurrency analysis through improved efficiency and user experience.
\section{Discussion}

\subsection{Transaction Execution}
A notable design choice in \toolname is the deliberate omission of a direct transaction execution module, a decision rooted in significant security considerations~\cite{zhang2019security,stephen2018review,leng2020blockchain}. We strictly define \toolname's role as a decision-support system, not an autonomous execution agent. While the AI demonstrates robust planning capabilities, the inherent risk of misinterpretation in complex financial commands, combined with the irreversibility of blockchain transactions~\cite{chen2024identifying,chen2025chatgpt}, presents an unacceptable potential for irrecoverable financial loss. An erroneous trade, once executed by an AI, cannot be undone. Therefore, to ensure a responsible and safe framework, the final transactional authority remains with the user. We plan to incorporate this feature into our system once more reliable and secure methods for automated on-chain interactions become available.

\subsection{Threats to Validity}
\noindent \textbf{External Threats.} A primary threat to validity concerns our system's dependence on underlying language models, representing an external threat beyond our direct control. The performance limitations, potential biases, or evolving capabilities of the primary LLM could significantly impact \toolname's effectiveness and reliability. To mitigate this concern, we designed our framework with flexible model selection, allowing seamless integration of newer, more capable models as they become available. 

\noindent \textbf{Internal Threats.} The primary internal threat stems from the subjectivity in our reward mechanism, where the judge model's semantic quality assessments could introduce evaluation bias. We addressed this by implementing a hybrid reward function that combines objective syntactic correctness checks with semantic evaluation, and by carefully engineering the judge prompt with explicit evaluation criteria and few-shot examples to ensure consistent scoring across different query types (refer to Appendix~\ref{app:judge_prompt}). 
\section{Conclusion}
In this paper, we propose \toolname, an intelligent agent system that enhances cryptocurrency investment through reinforcement learning–based tool selection and multi-agent collaboration. By addressing the limitations of static LLM agents and fragmented data platforms, \toolname enables adaptive reasoning, deeper analysis, and more reliable decision support. Experimental results show that the caller model achieves a 40.7\% improvement in recall and a 26.6\% increase in F1 score compared with the base model. User studies with 20 participants report average satisfaction with all the functionalities (4.64/5) and a clear preference over existing general LLMs or platforms (4.62/5). Overall, \toolname offers an effective and user-friendly solution for intelligent cryptocurrency investment.

\bibliographystyle{ACM-Reference-Format}
\bibliography{refs}


\begin{thebibliography}{50}


\ifx \showCODEN    \undefined \def \showCODEN     #1{\unskip}     \fi
\ifx \showISBNx    \undefined \def \showISBNx     #1{\unskip}     \fi
\ifx \showISBNxiii \undefined \def \showISBNxiii  #1{\unskip}     \fi
\ifx \showISSN     \undefined \def \showISSN      #1{\unskip}     \fi
\ifx \showLCCN     \undefined \def \showLCCN      #1{\unskip}     \fi
\ifx \shownote     \undefined \def \shownote      #1{#1}          \fi
\ifx \showarticletitle \undefined \def \showarticletitle #1{#1}   \fi
\ifx \showURL      \undefined \def \showURL       {\relax}        \fi
\providecommand\bibfield[2]{#2}
\providecommand\bibinfo[2]{#2}
\providecommand\natexlab[1]{#1}
\providecommand\showeprint[2][]{arXiv:#2}

\bibitem[bin(2025)]%
        {binance}
 \bibinfo{year}{2025}\natexlab{}.
\newblock \bibinfo{title}{Binance}.
\newblock \bibinfo{howpublished}{\url{https://www.binance.com}}.
\newblock


\bibitem[coi(2025a)]%
        {coindesk}
 \bibinfo{year}{2025}\natexlab{a}.
\newblock \bibinfo{title}{CoinDesk}.
\newblock \bibinfo{howpublished}{\url{https://www.coindesk.com/}}.
\newblock


\bibitem[coi(2025b)]%
        {coingecko}
 \bibinfo{year}{2025}\natexlab{b}.
\newblock \bibinfo{title}{CoinGecko}.
\newblock \bibinfo{howpublished}{\url{https://www.coingecko.com/}}.
\newblock


\bibitem[eth(2025)]%
        {etherscan}
 \bibinfo{year}{2025}\natexlab{}.
\newblock \bibinfo{title}{Etherscan}.
\newblock \bibinfo{howpublished}{\url{https://etherscan.io/}}.
\newblock


\bibitem[git(2025)]%
        {github}
 \bibinfo{year}{2025}\natexlab{}.
\newblock \bibinfo{title}{Github}.
\newblock \bibinfo{howpublished}{\url{https://github.com/}}.
\newblock


\bibitem[web(2025)]%
        {websearch}
 \bibinfo{year}{2025}\natexlab{}.
\newblock \bibinfo{title}{Introducing ChatGPT search}.
\newblock \bibinfo{howpublished}{\url{https://openai.com/index/introducing-chatgpt-search/}}.
\newblock


\bibitem[Ref(2025)]%
        {RefinitivEikon}
 \bibinfo{year}{2025}\natexlab{}.
\newblock \bibinfo{title}{Refinitiv Eikon}.
\newblock \bibinfo{howpublished}{\url{https://eikon.refinitiv.com/}}.
\newblock


\bibitem[sol(2025)]%
        {solidity}
 \bibinfo{year}{2025}\natexlab{}.
\newblock \bibinfo{title}{Solidity}.
\newblock \bibinfo{howpublished}{\url{https://docs.soliditylang.org/en/v0.8.30/}}.
\newblock


\bibitem[Achiam et~al\mbox{.}(2023)]%
        {achiam2023gpt}
\bibfield{author}{\bibinfo{person}{Josh Achiam}, \bibinfo{person}{Steven Adler}, \bibinfo{person}{Sandhini Agarwal}, \bibinfo{person}{Lama Ahmad}, \bibinfo{person}{Ilge Akkaya}, \bibinfo{person}{Florencia~Leoni Aleman}, \bibinfo{person}{Diogo Almeida}, \bibinfo{person}{Janko Altenschmidt}, \bibinfo{person}{Sam Altman}, \bibinfo{person}{Shyamal Anadkat}, {et~al\mbox{.}}} \bibinfo{year}{2023}\natexlab{}.
\newblock \showarticletitle{Gpt-4 technical report}.
\newblock \bibinfo{journal}{\emph{arXiv preprint arXiv:2303.08774}} (\bibinfo{year}{2023}).
\newblock


\bibitem[Almeida and Gon{\c{c}}alves(2022)]%
        {almeida2022systematic}
\bibfield{author}{\bibinfo{person}{Jos{\'e} Almeida} {and} \bibinfo{person}{Tiago~Cruz Gon{\c{c}}alves}.} \bibinfo{year}{2022}\natexlab{}.
\newblock \showarticletitle{A systematic literature review of volatility and risk management on cryptocurrency investment: A methodological point of view}.
\newblock \bibinfo{journal}{\emph{Risks}} \bibinfo{volume}{10}, \bibinfo{number}{5} (\bibinfo{year}{2022}), \bibinfo{pages}{107}.
\newblock


\bibitem[Chen et~al\mbox{.}(2025)]%
        {chen2025chatgpt}
\bibfield{author}{\bibinfo{person}{Chong Chen}, \bibinfo{person}{Jianzhong Su}, \bibinfo{person}{Jiachi Chen}, \bibinfo{person}{Yanlin Wang}, \bibinfo{person}{Tingting Bi}, \bibinfo{person}{Jianxing Yu}, \bibinfo{person}{Yanli Wang}, \bibinfo{person}{Xingwei Lin}, \bibinfo{person}{Ting Chen}, {and} \bibinfo{person}{Zibin Zheng}.} \bibinfo{year}{2025}\natexlab{}.
\newblock \showarticletitle{When chatgpt meets smart contract vulnerability detection: How far are we?}
\newblock \bibinfo{journal}{\emph{ACM Transactions on Software Engineering and Methodology}} \bibinfo{volume}{34}, \bibinfo{number}{4} (\bibinfo{year}{2025}), \bibinfo{pages}{1--30}.
\newblock


\bibitem[Chen et~al\mbox{.}(2024)]%
        {chen2024identifying}
\bibfield{author}{\bibinfo{person}{Jiachi Chen}, \bibinfo{person}{Chong Chen}, \bibinfo{person}{Jiang Hu}, \bibinfo{person}{John Grundy}, \bibinfo{person}{Yanlin Wang}, \bibinfo{person}{Ting Chen}, {and} \bibinfo{person}{Zibin Zheng}.} \bibinfo{year}{2024}\natexlab{}.
\newblock \showarticletitle{Identifying smart contract security issues in code snippets from stack overflow}. In \bibinfo{booktitle}{\emph{Proceedings of the 33rd ACM SIGSOFT International Symposium on Software Testing and Analysis}}. \bibinfo{pages}{1198--1210}.
\newblock


\bibitem[Feist et~al\mbox{.}(2019)]%
        {feist2019slither}
\bibfield{author}{\bibinfo{person}{Josselin Feist}, \bibinfo{person}{Gustavo Grieco}, {and} \bibinfo{person}{Alex Groce}.} \bibinfo{year}{2019}\natexlab{}.
\newblock \showarticletitle{Slither: a static analysis framework for smart contracts}. In \bibinfo{booktitle}{\emph{2019 IEEE/ACM 2nd International Workshop on Emerging Trends in Software Engineering for Blockchain (WETSEB)}}. IEEE, \bibinfo{pages}{8--15}.
\newblock


\bibitem[Gao et~al\mbox{.}(2023)]%
        {gao2023retrieval}
\bibfield{author}{\bibinfo{person}{Yunfan Gao}, \bibinfo{person}{Yun Xiong}, \bibinfo{person}{Xinyu Gao}, \bibinfo{person}{Kangxiang Jia}, \bibinfo{person}{Jinliu Pan}, \bibinfo{person}{Yuxi Bi}, \bibinfo{person}{Yixin Dai}, \bibinfo{person}{Jiawei Sun}, \bibinfo{person}{Haofen Wang}, {and} \bibinfo{person}{Haofen Wang}.} \bibinfo{year}{2023}\natexlab{}.
\newblock \showarticletitle{Retrieval-augmented generation for large language models: A survey}.
\newblock \bibinfo{journal}{\emph{arXiv preprint arXiv:2312.10997}} \bibinfo{volume}{2}, \bibinfo{number}{1} (\bibinfo{year}{2023}).
\newblock


\bibitem[Google(2025)]%
        {gemini}
\bibfield{author}{\bibinfo{person}{Google}.} \bibinfo{year}{2025}\natexlab{}.
\newblock \bibinfo{title}{Gemini}.
\newblock \bibinfo{howpublished}{\url{https://gemini.google.com/app}}.
\newblock


\bibitem[Grattafiori et~al\mbox{.}(2024)]%
        {grattafiori2024llama}
\bibfield{author}{\bibinfo{person}{Aaron Grattafiori}, \bibinfo{person}{Abhimanyu Dubey}, \bibinfo{person}{Abhinav Jauhri}, \bibinfo{person}{Abhinav Pandey}, \bibinfo{person}{Abhishek Kadian}, \bibinfo{person}{Ahmad Al-Dahle}, \bibinfo{person}{Aiesha Letman}, \bibinfo{person}{Akhil Mathur}, \bibinfo{person}{Alan Schelten}, \bibinfo{person}{Alex Vaughan}, {et~al\mbox{.}}} \bibinfo{year}{2024}\natexlab{}.
\newblock \showarticletitle{The llama 3 herd of models}.
\newblock \bibinfo{journal}{\emph{arXiv preprint arXiv:2407.21783}} (\bibinfo{year}{2024}).
\newblock


\bibitem[H{\"a}rdle et~al\mbox{.}(2020)]%
        {hardle2020understanding}
\bibfield{author}{\bibinfo{person}{Wolfgang~Karl H{\"a}rdle}, \bibinfo{person}{Campbell~R Harvey}, {and} \bibinfo{person}{Raphael~CG Reule}.} \bibinfo{year}{2020}\natexlab{}.
\newblock \bibinfo{title}{Understanding cryptocurrencies}.
\newblock \bibinfo{numpages}{181--208}~pages.
\newblock


\bibitem[Hu et~al\mbox{.}(2022)]%
        {hu2022lora}
\bibfield{author}{\bibinfo{person}{Edward~J Hu}, \bibinfo{person}{Yelong Shen}, \bibinfo{person}{Phillip Wallis}, \bibinfo{person}{Zeyuan Allen-Zhu}, \bibinfo{person}{Yuanzhi Li}, \bibinfo{person}{Shean Wang}, \bibinfo{person}{Lu Wang}, \bibinfo{person}{Weizhu Chen}, {et~al\mbox{.}}} \bibinfo{year}{2022}\natexlab{}.
\newblock \showarticletitle{Lora: Low-rank adaptation of large language models.}
\newblock \bibinfo{journal}{\emph{ICLR}} \bibinfo{volume}{1}, \bibinfo{number}{2} (\bibinfo{year}{2022}), \bibinfo{pages}{3}.
\newblock


\bibitem[Inci and Lagasse(2019)]%
        {inci2019cryptocurrencies}
\bibfield{author}{\bibinfo{person}{A~Can Inci} {and} \bibinfo{person}{Rachel Lagasse}.} \bibinfo{year}{2019}\natexlab{}.
\newblock \showarticletitle{Cryptocurrencies: applications and investment opportunities}.
\newblock \bibinfo{journal}{\emph{Journal of Capital Markets Studies}} \bibinfo{volume}{3}, \bibinfo{number}{2} (\bibinfo{year}{2019}), \bibinfo{pages}{98--112}.
\newblock


\bibitem[Joshi et~al\mbox{.}(2015)]%
        {joshi2015likert}
\bibfield{author}{\bibinfo{person}{Ankur Joshi}, \bibinfo{person}{Saket Kale}, \bibinfo{person}{Satish Chandel}, {and} \bibinfo{person}{D~Kumar Pal}.} \bibinfo{year}{2015}\natexlab{}.
\newblock \showarticletitle{Likert scale: Explored and explained}.
\newblock \bibinfo{journal}{\emph{British journal of applied science \& technology}} \bibinfo{volume}{7}, \bibinfo{number}{4} (\bibinfo{year}{2015}), \bibinfo{pages}{396}.
\newblock


\bibitem[Lee et~al\mbox{.}(2017)]%
        {lee2017cryptocurrency}
\bibfield{author}{\bibinfo{person}{David Kuo~Chuen Lee}, \bibinfo{person}{Li Guo}, {and} \bibinfo{person}{Yu Wang}.} \bibinfo{year}{2017}\natexlab{}.
\newblock \showarticletitle{Cryptocurrency: A new investment opportunity?}
\newblock \bibinfo{journal}{\emph{Available at SSRN 2994097}} (\bibinfo{year}{2017}).
\newblock


\bibitem[Leng et~al\mbox{.}(2020)]%
        {leng2020blockchain}
\bibfield{author}{\bibinfo{person}{Jiewu Leng}, \bibinfo{person}{Man Zhou}, \bibinfo{person}{J~Leon Zhao}, \bibinfo{person}{Yongfeng Huang}, {and} \bibinfo{person}{Yiyang Bian}.} \bibinfo{year}{2020}\natexlab{}.
\newblock \showarticletitle{Blockchain security: A survey of techniques and research directions}.
\newblock \bibinfo{journal}{\emph{IEEE Transactions on Services Computing}} \bibinfo{volume}{15}, \bibinfo{number}{4} (\bibinfo{year}{2020}), \bibinfo{pages}{2490--2510}.
\newblock


\bibitem[Lewis et~al\mbox{.}(2020)]%
        {lewis2020retrieval}
\bibfield{author}{\bibinfo{person}{Patrick Lewis}, \bibinfo{person}{Ethan Perez}, \bibinfo{person}{Aleksandra Piktus}, \bibinfo{person}{Fabio Petroni}, \bibinfo{person}{Vladimir Karpukhin}, \bibinfo{person}{Naman Goyal}, \bibinfo{person}{Heinrich K{\"u}ttler}, \bibinfo{person}{Mike Lewis}, \bibinfo{person}{Wen-tau Yih}, \bibinfo{person}{Tim Rockt{\"a}schel}, {et~al\mbox{.}}} \bibinfo{year}{2020}\natexlab{}.
\newblock \showarticletitle{Retrieval-augmented generation for knowledge-intensive nlp tasks}.
\newblock \bibinfo{journal}{\emph{Advances in neural information processing systems}}  \bibinfo{volume}{33} (\bibinfo{year}{2020}), \bibinfo{pages}{9459--9474}.
\newblock


\bibitem[Liu et~al\mbox{.}(2024)]%
        {liu2024deepseek}
\bibfield{author}{\bibinfo{person}{Aixin Liu}, \bibinfo{person}{Bei Feng}, \bibinfo{person}{Bing Xue}, \bibinfo{person}{Bingxuan Wang}, \bibinfo{person}{Bochao Wu}, \bibinfo{person}{Chengda Lu}, \bibinfo{person}{Chenggang Zhao}, \bibinfo{person}{Chengqi Deng}, \bibinfo{person}{Chenyu Zhang}, \bibinfo{person}{Chong Ruan}, {et~al\mbox{.}}} \bibinfo{year}{2024}\natexlab{}.
\newblock \showarticletitle{Deepseek-v3 technical report}.
\newblock \bibinfo{journal}{\emph{arXiv preprint arXiv:2412.19437}} (\bibinfo{year}{2024}).
\newblock


\bibitem[Luo et~al\mbox{.}(2025)]%
        {luo2025llm}
\bibfield{author}{\bibinfo{person}{Yichen Luo}, \bibinfo{person}{Yebo Feng}, \bibinfo{person}{Jiahua Xu}, \bibinfo{person}{Paolo Tasca}, {and} \bibinfo{person}{Yang Liu}.} \bibinfo{year}{2025}\natexlab{}.
\newblock \showarticletitle{LLM-Powered Multi-Agent System for Automated Crypto Portfolio Management}.
\newblock \bibinfo{journal}{\emph{arXiv preprint arXiv:2501.00826}} (\bibinfo{year}{2025}).
\newblock


\bibitem[Morck et~al\mbox{.}(1990)]%
        {morck1990stock}
\bibfield{author}{\bibinfo{person}{Randall Morck}, \bibinfo{person}{Andrei Shleifer}, \bibinfo{person}{Robert~W Vishny}, \bibinfo{person}{Matthew Shapiro}, {and} \bibinfo{person}{James~M Poterba}.} \bibinfo{year}{1990}\natexlab{}.
\newblock \showarticletitle{The stock market and investment: is the market a sideshow?}
\newblock \bibinfo{journal}{\emph{Brookings papers on economic Activity}} \bibinfo{volume}{1990}, \bibinfo{number}{2} (\bibinfo{year}{1990}), \bibinfo{pages}{157--215}.
\newblock


\bibitem[OpenAI(2025)]%
        {gpt4o}
\bibfield{author}{\bibinfo{person}{OpenAI}.} \bibinfo{year}{2025}\natexlab{}.
\newblock \bibinfo{title}{GPT-4o}.
\newblock \bibinfo{howpublished}{\url{https://platform.openai.com/docs/models/gpt-4o}}.
\newblock


\bibitem[Patil et~al\mbox{.}(2025)]%
        {patil2025bfcl}
\bibfield{author}{\bibinfo{person}{Shishir~G. Patil}, \bibinfo{person}{Huanzhi Mao}, \bibinfo{person}{Charlie Cheng-Jie~Ji}, \bibinfo{person}{Fanjia Yan}, \bibinfo{person}{Vishnu Suresh}, \bibinfo{person}{Ion Stoica}, {and} \bibinfo{person}{Joseph E.~Gonzalez}.} \bibinfo{year}{2025}\natexlab{}.
\newblock \showarticletitle{The Berkeley Function Calling Leaderboard (BFCL): From Tool Use to Agentic Evaluation of Large Language Models}. In \bibinfo{booktitle}{\emph{Forty-second International Conference on Machine Learning}}.
\newblock


\bibitem[Petrov and Monroy-Hern{\'a}ndez(2023)]%
        {petrov2023dream}
\bibfield{author}{\bibinfo{person}{Elizabeth Petrov} {and} \bibinfo{person}{Andr{\'e}s Monroy-Hern{\'a}ndez}.} \bibinfo{year}{2023}\natexlab{}.
\newblock \showarticletitle{Dream Garden: Exploring Location-Based, Collaboratively-Created Augmented Reality Spaces}. In \bibinfo{booktitle}{\emph{Extended Abstracts of the 2023 CHI Conference on Human Factors in Computing Systems}}. \bibinfo{pages}{1--6}.
\newblock


\bibitem[Phillip et~al\mbox{.}(2018)]%
        {phillip2018new}
\bibfield{author}{\bibinfo{person}{Andrew Phillip}, \bibinfo{person}{Jennifer~SK Chan}, {and} \bibinfo{person}{Shelton Peiris}.} \bibinfo{year}{2018}\natexlab{}.
\newblock \showarticletitle{A new look at cryptocurrencies}.
\newblock \bibinfo{journal}{\emph{Economics letters}}  \bibinfo{volume}{163} (\bibinfo{year}{2018}), \bibinfo{pages}{6--9}.
\newblock


\bibitem[Prabhakar et~al\mbox{.}(2025)]%
        {prabhakar2025apigen}
\bibfield{author}{\bibinfo{person}{Akshara Prabhakar}, \bibinfo{person}{Zuxin Liu}, \bibinfo{person}{Ming Zhu}, \bibinfo{person}{Jianguo Zhang}, \bibinfo{person}{Tulika Awalgaonkar}, \bibinfo{person}{Shiyu Wang}, \bibinfo{person}{Zhiwei Liu}, \bibinfo{person}{Haolin Chen}, \bibinfo{person}{Thai Hoang}, {et~al\mbox{.}}} \bibinfo{year}{2025}\natexlab{}.
\newblock \showarticletitle{APIGen-MT: Agentic PIpeline for Multi-Turn Data Generation via Simulated Agent-Human Interplay}.
\newblock \bibinfo{journal}{\emph{arXiv preprint arXiv:2504.03601}} (\bibinfo{year}{2025}).
\newblock


\bibitem[Puterman(1990)]%
        {puterman1990markov}
\bibfield{author}{\bibinfo{person}{Martin~L Puterman}.} \bibinfo{year}{1990}\natexlab{}.
\newblock \showarticletitle{Markov decision processes}.
\newblock \bibinfo{journal}{\emph{Handbooks in operations research and management science}}  \bibinfo{volume}{2} (\bibinfo{year}{1990}), \bibinfo{pages}{331--434}.
\newblock


\bibitem[Ross et~al\mbox{.}(2025)]%
        {ross2025when2call}
\bibfield{author}{\bibinfo{person}{Hayley Ross}, \bibinfo{person}{Ameya~Sunil Mahabaleshwarkar}, {and} \bibinfo{person}{Yoshi Suhara}.} \bibinfo{year}{2025}\natexlab{}.
\newblock \showarticletitle{When2Call: When (not) to Call Tools}.
\newblock \bibinfo{journal}{\emph{arXiv preprint arXiv:2504.18851}} (\bibinfo{year}{2025}).
\newblock


\bibitem[Schulman et~al\mbox{.}(2017)]%
        {schulman2017proximal}
\bibfield{author}{\bibinfo{person}{John Schulman}, \bibinfo{person}{Filip Wolski}, \bibinfo{person}{Prafulla Dhariwal}, \bibinfo{person}{Alec Radford}, {and} \bibinfo{person}{Oleg Klimov}.} \bibinfo{year}{2017}\natexlab{}.
\newblock \showarticletitle{Proximal policy optimization algorithms}.
\newblock \bibinfo{journal}{\emph{arXiv preprint arXiv:1707.06347}} (\bibinfo{year}{2017}).
\newblock


\bibitem[Shen(2024)]%
        {shen2024llm}
\bibfield{author}{\bibinfo{person}{Zhuocheng Shen}.} \bibinfo{year}{2024}\natexlab{}.
\newblock \showarticletitle{Llm with tools: A survey}.
\newblock \bibinfo{journal}{\emph{arXiv preprint arXiv:2409.18807}} (\bibinfo{year}{2024}).
\newblock


\bibitem[Stephen and Alex(2018)]%
        {stephen2018review}
\bibfield{author}{\bibinfo{person}{Remya Stephen} {and} \bibinfo{person}{Aneena Alex}.} \bibinfo{year}{2018}\natexlab{}.
\newblock \showarticletitle{A review on blockchain security}. In \bibinfo{booktitle}{\emph{IOP conference series: materials science and engineering}}, Vol.~\bibinfo{volume}{396}. IOP Publishing, \bibinfo{pages}{012030}.
\newblock


\bibitem[Team(2025a)]%
        {qwen3}
\bibfield{author}{\bibinfo{person}{Qwen Team}.} \bibinfo{year}{2025}\natexlab{a}.
\newblock \bibinfo{title}{Qwen3}.
\newblock \bibinfo{howpublished}{\url{https://qwenlm.github.io/blog/qwen3/}}.
\newblock


\bibitem[Team(2025b)]%
        {qwen3technicalreport}
\bibfield{author}{\bibinfo{person}{Qwen Team}.} \bibinfo{year}{2025}\natexlab{b}.
\newblock \bibinfo{title}{Qwen3 Technical Report}.
\newblock
\showeprint[arxiv]{2505.09388}~[cs.CL]
\urldef\tempurl%
\url{https://arxiv.org/abs/2505.09388}
\showURL{%
\tempurl}


\bibitem[von Werra et~al\mbox{.}(2020)]%
        {vonwerra2022trl}
\bibfield{author}{\bibinfo{person}{Leandro von Werra}, \bibinfo{person}{Younes Belkada}, \bibinfo{person}{Lewis Tunstall}, \bibinfo{person}{Edward Beeching}, \bibinfo{person}{Tristan Thrush}, \bibinfo{person}{Nathan Lambert}, \bibinfo{person}{Shengyi Huang}, \bibinfo{person}{Kashif Rasul}, {and} \bibinfo{person}{Quentin Gallouédec}.} \bibinfo{year}{2020}\natexlab{}.
\newblock \bibinfo{title}{TRL: Transformer Reinforcement Learning}.
\newblock \bibinfo{howpublished}{\url{https://github.com/huggingface/trl}}.
\newblock


\bibitem[Wang et~al\mbox{.}(2024)]%
        {wang2024toolflow}
\bibfield{author}{\bibinfo{person}{Zezhong Wang}, \bibinfo{person}{Xingshan Zeng}, \bibinfo{person}{Weiwen Liu}, \bibinfo{person}{Liangyou Li}, \bibinfo{person}{Yasheng Wang}, \bibinfo{person}{Lifeng Shang}, \bibinfo{person}{Xin Jiang}, \bibinfo{person}{Qun Liu}, {and} \bibinfo{person}{Kam-Fai Wong}.} \bibinfo{year}{2024}\natexlab{}.
\newblock \showarticletitle{Toolflow: Boosting llm tool-calling through natural and coherent dialogue synthesis}.
\newblock \bibinfo{journal}{\emph{arXiv preprint arXiv:2410.18447}} (\bibinfo{year}{2024}).
\newblock


\bibitem[Xiao et~al\mbox{.}(2024)]%
        {xiao2024tradingagents}
\bibfield{author}{\bibinfo{person}{Yijia Xiao}, \bibinfo{person}{Edward Sun}, \bibinfo{person}{Di Luo}, {and} \bibinfo{person}{Wei Wang}.} \bibinfo{year}{2024}\natexlab{}.
\newblock \showarticletitle{TradingAgents: Multi-agents LLM financial trading framework}.
\newblock \bibinfo{journal}{\emph{arXiv preprint arXiv:2412.20138}} (\bibinfo{year}{2024}).
\newblock


\bibitem[Xu et~al\mbox{.}(2024)]%
        {xu2024reducing}
\bibfield{author}{\bibinfo{person}{Hongshen Xu}, \bibinfo{person}{Zichen Zhu}, \bibinfo{person}{Lei Pan}, \bibinfo{person}{Zihan Wang}, \bibinfo{person}{Su Zhu}, \bibinfo{person}{Da Ma}, \bibinfo{person}{Ruisheng Cao}, \bibinfo{person}{Lu Chen}, {and} \bibinfo{person}{Kai Yu}.} \bibinfo{year}{2024}\natexlab{}.
\newblock \showarticletitle{Reducing tool hallucination via reliability alignment}.
\newblock \bibinfo{journal}{\emph{arXiv preprint arXiv:2412.04141}} (\bibinfo{year}{2024}).
\newblock


\bibitem[Yu(2022)]%
        {yu2022retrieval}
\bibfield{author}{\bibinfo{person}{Wenhao Yu}.} \bibinfo{year}{2022}\natexlab{}.
\newblock \showarticletitle{Retrieval-augmented generation across heterogeneous knowledge}. In \bibinfo{booktitle}{\emph{Proceedings of the 2022 conference of the North American chapter of the association for computational linguistics: human language technologies: student research workshop}}. \bibinfo{pages}{52--58}.
\newblock


\bibitem[Yu et~al\mbox{.}(2024)]%
        {yu2024fincon}
\bibfield{author}{\bibinfo{person}{Yangyang Yu}, \bibinfo{person}{Zhiyuan Yao}, \bibinfo{person}{Haohang Li}, \bibinfo{person}{Zhiyang Deng}, \bibinfo{person}{Yuechen Jiang}, \bibinfo{person}{Yupeng Cao}, \bibinfo{person}{Zhi Chen}, \bibinfo{person}{Jordan Suchow}, \bibinfo{person}{Zhenyu Cui}, \bibinfo{person}{Rong Liu}, {et~al\mbox{.}}} \bibinfo{year}{2024}\natexlab{}.
\newblock \showarticletitle{Fincon: A synthesized llm multi-agent system with conceptual verbal reinforcement for enhanced financial decision making}.
\newblock \bibinfo{journal}{\emph{Advances in Neural Information Processing Systems}}  \bibinfo{volume}{37} (\bibinfo{year}{2024}), \bibinfo{pages}{137010--137045}.
\newblock


\bibitem[Yuan et~al\mbox{.}(2024)]%
        {yuan2024easytool}
\bibfield{author}{\bibinfo{person}{Siyu Yuan}, \bibinfo{person}{Kaitao Song}, \bibinfo{person}{Jiangjie Chen}, \bibinfo{person}{Xu Tan}, \bibinfo{person}{Yongliang Shen}, \bibinfo{person}{Ren Kan}, \bibinfo{person}{Dongsheng Li}, {and} \bibinfo{person}{Deqing Yang}.} \bibinfo{year}{2024}\natexlab{}.
\newblock \showarticletitle{Easytool: Enhancing llm-based agents with concise tool instruction}.
\newblock \bibinfo{journal}{\emph{arXiv preprint arXiv:2401.06201}} (\bibinfo{year}{2024}).
\newblock


\bibitem[Yuniningsih et~al\mbox{.}(2017)]%
        {yuniningsih2017analysis}
\bibfield{author}{\bibinfo{person}{Yuniningsih Yuniningsih}, \bibinfo{person}{Sugeng Widodo}, {and} \bibinfo{person}{Muh Barid~Nizarudin Wajdi}.} \bibinfo{year}{2017}\natexlab{}.
\newblock \showarticletitle{An analysis of decision making in the stock investment}.
\newblock \bibinfo{journal}{\emph{Economic: Journal of Economic and Islamic Law}} \bibinfo{volume}{8}, \bibinfo{number}{2} (\bibinfo{year}{2017}), \bibinfo{pages}{122--128}.
\newblock


\bibitem[Zhang et~al\mbox{.}(2019)]%
        {zhang2019security}
\bibfield{author}{\bibinfo{person}{Rui Zhang}, \bibinfo{person}{Rui Xue}, {and} \bibinfo{person}{Ling Liu}.} \bibinfo{year}{2019}\natexlab{}.
\newblock \showarticletitle{Security and privacy on blockchain}.
\newblock \bibinfo{journal}{\emph{ACM Computing Surveys (CSUR)}} \bibinfo{volume}{52}, \bibinfo{number}{3} (\bibinfo{year}{2019}), \bibinfo{pages}{1--34}.
\newblock


\bibitem[Zhang et~al\mbox{.}(2025)]%
        {zhang2025prompting}
\bibfield{author}{\bibinfo{person}{Tianyi Zhang}, \bibinfo{person}{Colin Au~Yeung}, \bibinfo{person}{Emily Aurelia}, \bibinfo{person}{Yuki Onishi}, \bibinfo{person}{Neil Chulpongsatorn}, \bibinfo{person}{Jiannan Li}, {and} \bibinfo{person}{Anthony Tang}.} \bibinfo{year}{2025}\natexlab{}.
\newblock \showarticletitle{Prompting an Embodied AI Agent: How Embodiment and Multimodal Signaling Affects Prompting Behaviour}. In \bibinfo{booktitle}{\emph{Proceedings of the 2025 CHI Conference on Human Factors in Computing Systems}}. \bibinfo{pages}{1--25}.
\newblock


\bibitem[Zhao and Zhang(2021)]%
        {zhao2021financial}
\bibfield{author}{\bibinfo{person}{Haidong Zhao} {and} \bibinfo{person}{Lini Zhang}.} \bibinfo{year}{2021}\natexlab{}.
\newblock \showarticletitle{Financial literacy or investment experience: which is more influential in cryptocurrency investment?}
\newblock \bibinfo{journal}{\emph{International Journal of Bank Marketing}} \bibinfo{volume}{39}, \bibinfo{number}{7} (\bibinfo{year}{2021}), \bibinfo{pages}{1208--1226}.
\newblock


\bibitem[Zhao et~al\mbox{.}(2024)]%
        {zhao2024retrieval}
\bibfield{author}{\bibinfo{person}{Penghao Zhao}, \bibinfo{person}{Hailin Zhang}, \bibinfo{person}{Qinhan Yu}, \bibinfo{person}{Zhengren Wang}, \bibinfo{person}{Yunteng Geng}, \bibinfo{person}{Fangcheng Fu}, \bibinfo{person}{Ling Yang}, \bibinfo{person}{Wentao Zhang}, \bibinfo{person}{Jie Jiang}, {and} \bibinfo{person}{Bin Cui}.} \bibinfo{year}{2024}\natexlab{}.
\newblock \showarticletitle{Retrieval-augmented generation for ai-generated content: A survey}.
\newblock \bibinfo{journal}{\emph{arXiv preprint arXiv:2402.19473}} (\bibinfo{year}{2024}).
\newblock


\end{thebibliography}

\appendix
\section{Appendix}

\subsection{Prompt of Judge Model}
\label{app:judge_prompt}

The judge model plays a critical role in the reinforcement learning pipeline by providing nuanced semantic evaluation of the agent's tool-calling strategies. This section details the prompt engineering methodology employed to ensure consistent and reliable feedback signals.

\subsubsection{Design Principles}
The judge prompt was designed following several key principles:

\noindent \textbf{Explicit Role Definition.} The prompt begins by clearly establishing the judge's role as an evaluator of tool usage plans in the cryptocurrency investment domain. This contextual framing ensures the model understands it should assess tool selection from an investment analysis perspective.

\noindent \textbf{Comprehensive Tool Awareness.} The complete list of available tools with their descriptions and parameters is embedded in the prompt. This ensures the judge has full knowledge of the agent's action space and can accurately assess whether appropriate tools were selected or valuable tools were overlooked.

\noindent \textbf{Structured Evaluation Criteria.} Rather than requesting a holistic quality score, the prompt decomposes evaluation into two distinct dimensions, i.e., Information Coverage and Relevance, each with explicit definitions and scoring guidelines. This structured approach reduces ambiguity and improves scoring consistency across different query types.

\noindent \textbf{Few-Shot Calibration.} The prompt includes carefully curated examples that demonstrate the desired evaluation behavior across diverse scenarios, serving as anchors for the model's scoring scale.

\subsubsection{Evaluation Criteria Specification}

The prompt defines two evaluation dimensions with explicit scoring guidelines:

\noindent \textbf{Information Coverage (0-10):} This metric assesses the breadth and completeness of the investigation strategy. The prompt instructs the judge to consider:
\begin{itemize}
\item Whether multiple relevant information dimensions are addressed (e.g., market data, on-chain analysis, fundamental research, news sentiment)
\item If the selected tools can collectively provide sufficient context for an informed decision
\item Whether critical aspects of cryptocurrency investment analysis are missing from the plan
\item The comprehensiveness of the information gathering strategy relative to query complexity
\end{itemize}

The prompt explicitly encourages rewarding multi-faceted analysis approaches. For complex investment queries like ``Should I invest in Token X?'', a comprehensive plan should incorporate market analysis, project fundamentals, security assessment, and recent news—not merely fetch the current price. This design principle directly addresses the motivating problem that standard tool-augmented LLMs tend toward minimal tool usage.

\noindent \textbf{Relevance (0-10):} This metric evaluates the precision and appropriateness of tool selection. The prompt guides the judge to assess:
\begin{itemize}
\item Whether each selected tool is appropriate for the query context
\item If tool parameters are correctly specified and aligned with the user's intent
\item Whether any tools are redundant, misapplied, or tangential to the query
\end{itemize}

This dimension serves as a quality control mechanism, preventing the agent from invoking tools indiscriminately to maximize coverage scores. The prompt clarifies that comprehensive coverage should not come at the cost of relevance, which means every tool call should serve a clear purpose.

\subsubsection{Few-Shot Example Construction}

To calibrate the judge model's scoring behavior, the prompt includes three carefully designed few-shot examples:

\noindent \textbf{Example 1: High-Quality Comprehensive Plan}
\begin{itemize}
\item \textit{Query:} ``Is Bitcoin a good investment right now?''
\item \textit{Tool Calls:} [get\_crypto\_price, get\_recent\_news, kline\_analysis, \\project\_background\_agent, transaction\_analysis]
\item \textit{Expected Scores:} Coverage: 9.0, Relevance: 9.5
\item \textit{Rationale:} Demonstrates ideal behavior for complex investment queries—multiple complementary tools covering market data, news sentiment, technical analysis, fundamental background, and on-chain activity
\end{itemize}

\noindent \textbf{Example 2: Low-Quality Minimal Plan}
\begin{itemize}
\item \textit{Query:} ``Is Bitcoin a good investment right now?''
\item \textit{Tool Calls:} [get\_crypto\_price]
\item \textit{Expected Scores:} Coverage: 3.0, Relevance: 6.0
\item \textit{Rationale:} Illustrates insufficient analysis—while the price tool is relevant (hence moderate relevance score), it provides only one-dimensional information inadequate for investment decisions (hence low coverage score)
\end{itemize}

\noindent \textbf{Example 3: Correct Empty Plan}
\begin{itemize}
\item \textit{Query:} ``Can you explain what blockchain technology is?''
\item \textit{Tool Calls:} [] (empty list)
\item \textit{Expected Scores:} Coverage: 10.0, Relevance: 10.0
\item \textit{Rationale:} Demonstrates that no tools are needed for queries that can be answered from parametric knowledge—prevents penalizing the agent for correctly identifying when tools are unnecessary
\end{itemize}

The third example is particularly crucial. Without it, the judge might incorrectly penalize the agent for not using tools on queries that don't require external information, leading to inefficient over-invocation of tools. By explicitly showing that an empty tool list can receive perfect scores for appropriate queries, the prompt teaches the agent to be selective and efficient.

\subsubsection{Scoring Scale Anchoring}

The prompt provides explicit anchoring guidance for the 0-10 scale:
\begin{itemize}
\item \textbf{9-10:} Excellent—comprehensive coverage of all relevant dimensions (for coverage), or perfect precision with no unnecessary tools (for relevance)
\item \textbf{7-8:} Good—covers most important aspects with minor gaps (for coverage), or mostly relevant with minor issues (for relevance)
\item \textbf{5-6:} Adequate—covers some key aspects but missing important dimensions (for coverage), or generally relevant but with notable issues (for relevance)
\item \textbf{3-4:} Poor—significant gaps in coverage (for coverage), or contains irrelevant or misapplied tools (for relevance)
\item \textbf{0-2:} Very Poor—minimal (for coverage) or completely inappropriate information gathering (for relevance)
\end{itemize}

This anchoring helps maintain consistency across different evaluation sessions and reduces score inflation or deflation that might occur without explicit scale definitions.





\begin{figure}[h]
\setlength{\belowcaptionskip}{-0.5cm}
\centering
\includegraphics[width=\linewidth]{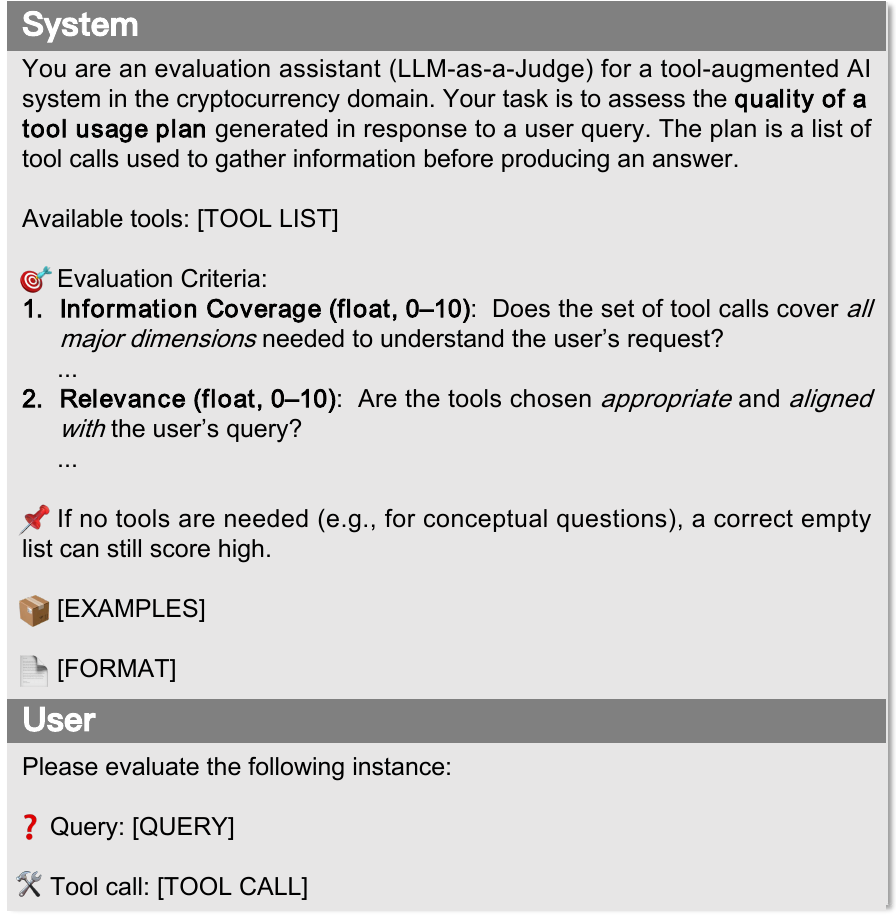}
\caption{Prompt template of judge model, including system prompt with role definition, evaluation criteria, few-shot examples, and output format specification.}
\label{fig:prompt}
\end{figure}

\subsection{Detailed Functional Module Descriptions}
\label{app:detailed_functions}

\begin{table*}[h]
\centering
\caption{Overview of \toolname's Functional Modules and Analytical Capabilities}
\label{tab:functions}
\begin{tabular}{p{5cm}|p{6cm}|p{5.5cm}}
\hline
\textbf{Functions} & \textbf{Data Retrieved} & \textbf{Advanced Analysis Capabilities} \\
\hline
\hline
Market Analysis & Real-time price, trading volume, price change rates, top gainers/losers and K-line chart data from Binance API & Price trend analysis, volatility assessment, market momentum evaluation \\
\hline
Transaction Analysis & Large-scale token transfers ($>$\$1,000,000) from Etherscan API, including sender/receiver addresses and timestamps & Intent inference (whale movements, exchange deposits), potential market manipulation detection \\
\hline
Smart Contract Security Analysis & Verified source code from Etherscan and vulnerability reports from Slither static analysis tool & Risk assessment through vulnerability pattern detection, false positive filtering, and security severity classification \\
\hline
Project Background Agent (PBAgent) & Project overview, team information, technical specifications from web sources, GitHub repositories, and whitepapers & Fundamental value assessment, tokenomics evaluation, and comprehensive project credibility analysis \\
\hline
Historical Events Agent (HEAgent) & Security incidents, regulatory announcements, and protocol upgrades from CoinDesk API & Cause-effect relationship identification, long-term risk pattern recognition, and event-driven market impact analysis \\
\hline
Crypto News Agent (CNAgent) & Real-time cryptocurrency market news and announcements from CoinDesk API & Market sentiment analysis, affected asset identification, and trend signal detection \\
\hline
\end{tabular}
\end{table*}

This section provides an in-depth description of each functional module presented in Table~\ref{tab:functions}, elaborating on their implementation, data processing pipelines, and analytical methodologies.

\noindent \textbf{Market Analysis Module.} The market analysis module serves as the primary interface for retrieving real-time market data. It integrates the Binance API~\cite{binance} to fetch current prices, 24-hour trading volumes, and price change percentages for specified cryptocurrencies. When users query about specific tokens, the module fetches current prices, trading volumes, price change percentages, and K-line (candlestick) chart data. The inclusion of visual K-line data introduces a multimodal dimension to the analysis, enabling the system to leverage multimodal models that analyze chart patterns based on economic principles and technical analysis theory. When investment targets are unspecified or users request general market conditions, the module switches to market overview mode, retrieving ranked lists of top gainers, losers, and high-volume assets.

\noindent \textbf{Transaction Analysis Module.} This module monitors on-chain activities to detect significant token movements that may signal market shifts. By querying the Etherscan~\cite{etherscan} API, it retrieves recent transactions exceeding \$1,000,000 in value. This threshold filters out retail trading noise to focus on whale movements with potential market impact. For each significant transaction, the module extracts transaction intent information directly from Etherscan's web interface, which provides labeled classifications such as transfers to exchange addresses (suggesting potential selling pressure), transfers from exchanges (indicating accumulation), and peer-to-peer transfers between large holders (signaling strategic repositioning). The module presents findings in a structured format, including transaction hash, timestamp, sender/receiver classification, and inferred market implications.

\noindent \textbf{Smart Contract Security Analysis Module.} Smart contract vulnerabilities pose significant existential risks to cryptocurrency investments. Unlike traditional financial instruments, code vulnerabilities can result in the complete and irreversible loss of invested capital. This module provides an automated security assessment to identify technical risks before investment.

\noindent \textbf{Multi-Stage Analysis Pipeline:} As illustrated in Figure~\ref{fig:code_analysis} in the main text, the module implements a sophisticated three-stage pipeline:

\textit{Stage 1: Source Code Retrieval}
\begin{itemize}
\item Queries Etherscan API~\cite{etherscan} using the contract address
\item Retrieves verified source code (Solidity~\cite{solidity}), including all imported dependencies
\item Supports multi-file contracts by reconstructing the complete codebase
\end{itemize}

\textit{Stage 2: Static Analysis with Slither}
\begin{itemize}
\item Invokes \textit{Slither}~\cite{feist2019slither}, a widely adopted static analysis framework for Solidity~\cite{solidity}
\item \textit{Slither}~\cite{feist2019slither} performs control-flow analysis, data-flow analysis, and pattern matching to detect 70+ vulnerability types
\item Raw output includes severity classification (high, medium, low), affected code locations, and technical descriptions
\end{itemize}

\textit{Stage 3: LLM-Based Refinement and Report Generation}
\begin{itemize}
\item \textbf{Filtering:} Only high and medium severity findings are retained; low-severity and informational issues are excluded to reduce noise
\item \textbf{Code Snippet Extraction:} For each vulnerability, the module extracts the surrounding code context to provide concrete examples
\item \textbf{False Positive Analysis:} An LLM analyzes each finding in context to identify potential false positives
\item \textbf{Structured Data Conversion:} Findings are converted to JSON format with fields: vulnerability type, severity, location (contract name, line number), description, code snippet
\item \textbf{Natural Language Report:} The LLM generates a security summary explaining vulnerabilities in plain language, their potential impact, and overall risk assessment
\end{itemize}

\noindent \textbf{Project Background Agent (PBAgent).} PBAgent conducts comprehensive fundamental analysis through three specialized components. \textit{ProjectInfoGatherer} uses web search to collect information from official websites, GitHub repositories, and whitepapers, capturing project mission, technical architecture, and development activity. \textit{ReportSynthesizer} structures unstructured content into standardized fields, including project overview, team background, tokenomics, and use cases. \textit{ProjectCritiquer} performs a critical evaluation, assessing token distribution fairness, technological innovation, and real-world utility, and compares the project against its competitors. This multi-stage process generates comprehensive fundamental analysis reports that help investors evaluate the long-term viability of projects.

\noindent \textbf{Historical Events Agent (HEAgent).} HEAgent provides temporal context by analyzing historical events that may have shaped current market conditions. \textit{EventsFetcher} retrieves dated event records from \textit{CoinDesk} API~\cite{coindesk}, including security incidents, regulatory announcements, protocol upgrades, and partnership announcements. \textit{EventsAnalyzer} examines these events to identify cause-and-effect relationships, assess long-term implications, and recognize recurring patterns in project responses to crises. This historical perspective helps investors understand a project's resilience, governance effectiveness, and potential future behavior under stress.

\noindent \textbf{Crypto News Agent (CNAgent).} CNAgent monitors real-time information flow to capture market-moving news. \textit{CryptoNewsFetcher} queries \textit{CoinDesk} API~\cite{coindesk} for the latest cryptocurrency news, filtered by relevance and recency. \textit{MarketImpactAnalyzer} performs sentiment analysis to classify news as bullish, bearish, or neutral, identifies which assets are directly or indirectly affected, and estimates the potential market impact magnitude based on the news's significance and market context. This enables proactive decision-making by alerting investors to emerging trends and potential market catalysts before they are fully reflected in market prices.

\vspace{-0.1cm}
\subsection{User Study Questionnaire Design}
\label{app:questionnaire_design}

The user study questionnaire was designed to systematically evaluate \toolname across multiple dimensions, capturing both quantitative assessments and qualitative insights from participants with diverse backgrounds in cryptocurrency investment.

\noindent \textbf{Participant Background Profiling (Q1-Q4).} Questions 1-4 establish participant context through role identification (nine categories including ordinary users, researchers, developers, and students), investment experience duration (less than 1 year to over 5 years), self-assessed expertise levels (novice, experienced, professional analyst), and existing information-seeking behaviors (multiple-choice covering news websites, social media, exchange analytics, and professional tools). This profiling enables response stratification across user segments and contextualizes subsequent evaluations within participants' existing workflows.

\noindent \textbf{Functional Module Evaluation (Q5-Q10).}
Questions 5-10 systematically evaluate each of \toolname's six core functional modules using matrix-style 5-point Likert scale items with N/A option. Each module assessment employs 3-4 tailored evaluation dimensions: transaction analysis (Q5) examines information accuracy and expression clarity; market analysis (Q6) assesses accuracy, analytical logic, and clarity; smart contract security (Q7) evaluates coverage, identification accuracy, and clarity; historical events (Q8) measures authenticity, comprehensiveness, description clarity, and analysis rationality; project background (Q9) assesses information coverage, accuracy, report clarity, and evaluation insight; crypto news (Q10) examines timeliness, relevance, summary accuracy, and analytical logic. This multi-dimensional decomposition enables fine-grained assessment of specific capabilities rather than holistic impressions.

\noindent \textbf{System-Level Reasoning Assessment (Q11).}
Question 11 evaluates the RL-enhanced caller model's tool orchestration performance through three matrix-style dimensions: selection rationality, sequence efficiency, and reasoning coherence. This directly assesses whether the reinforcement learning-based multi-step tool selection translates into a perceivable improvement in user experience.

\noindent \textbf{Comparative Analysis (Q12-Q13).}
Questions 12-13 position \toolname against current approaches. Q12 compares \toolname against general-purpose LLMs in terms of information integration capability, task understanding, reasoning clarity, and output practical utility. Q13 compares \toolname against cryptocurrency platforms in terms of information comprehensiveness, analytical depth, and quality of investment advice. These questions contextualize \toolname's performance within the existing tool ecosystem.

\noindent \textbf{Perceived Improvements and Overall Experience (Q14-Q17).}
Question 14 employs multiple-choice to identify improvements over traditional methods. Questions 15-17 are open-ended, soliciting qualitative feedback on the most satisfying aspects, encountered problems, areas for improvement, and additional suggestions. These capture nuanced perspectives and improvement priorities that structured questions might overlook, providing rich qualitative data to complement quantitative metrics.
\end{document}